\documentclass[11pt]{article}

\usepackage{acl}

\usepackage{times}
\usepackage{latexsym}

\usepackage[T1]{fontenc}

\usepackage[utf8]{inputenc}

\usepackage{microtype}

\usepackage{inconsolata}

\usepackage{graphicx}

\usepackage{amsmath}
\usepackage{amssymb}
\usepackage{mathtools}
\usepackage{amsthm}
\usepackage{booktabs}
\usepackage{multirow}
\usepackage{subcaption}
\usepackage{enumitem}

\title{TARPO: Token-Wise Latent-Explicit Reasoning via Action-Routing Policy Optimization}
\author{
  \textbf{Liting Zhang},
  \textbf{Shiwan Zhao},
  \textbf{Xuyang Zhao},
  \textbf{Zichen Xu},
  \textbf{Jianye Wang},
  \textbf{Qicheng Li\thanks{Corresponding author}}
  \\ \\
  TMCC, College of Computer Science, Nankai University, Tianjin, China \\
  \small{
    \textbf{Correspondence:} \href{mailto:zhangliting@mail.nankai.edu.cn}{zhangliting@mail.nankai.edu.cn},
    \href{mailto:liqicheng@nankai.edu.cn}{liqicheng@nankai.edu.cn}
  }
}

\begin{document}
\maketitle
\begin{abstract}
 Latent reasoning has emerged as a promising alternative to discrete Chain-of-Thought (CoT) in large language models (LLMs), enabling more expressive reasoning by operating over continuous representations. However, the inherently deterministic nature of continuous representations limits policy exploration in reinforcement learning (RL). To address this, we propose TARPO (Token-Wise Latent-Explicit Reasoning via Action-Routing Policy Optimization), a pure RL framework that adaptively switches between discrete token generation and continuous latent reasoning at each step. TARPO introduces a lightweight action head router that observes the current hidden state and samples a routing decision from a binary mode-selection space, preserving the stochasticity of discrete token sampling from the vocabulary. The LLM backbone and router are jointly optimized end-to-end with a shared group-relative advantage signal. Extensive experiments across Qwen2.5 (from 1.5B to 7B) and Llama-3.1-8B backbones demonstrate that TARPO consistently outperforms existing explicit and latent reasoning RL baselines across diverse benchmarks. Further analysis shows that TARPO learns adaptive token-wise switching behaviors while maintaining stable training dynamics. Our code is available at \url{https://github.com/NKU-LITI/TARPO-master}.


\end{abstract}




\section{Introduction}


Large Language Models (LLMs) have demonstrated remarkable reasoning capabilities under the Chain-of-Thought (CoT) paradigm~\citep{cot,deepseek-cot,kim2025-cot}. However, their reliance on discrete token generation introduces a fundamental bottleneck: high-dimensional hidden states must be collapsed into a single token at each step, limiting expressiveness and information capacity~\citep{coconut}. Latent reasoning has recently emerged as a promising paradigm~\citep{survey1,survey2}. By shifting reasoning from discrete tokens to continuous representations, it enables more expressive and efficient reasoning trajectories. 



Unlike discrete token sampling which is inherently stochastic, continuous representations are naturally deterministic, whether formulated as raw hidden states~\citep{coconut, system1.5} or probability-weighted embedding mixtures~\citep{soft-thinking}. This lack of native randomness fundamentally limits policy exploration in reinforcement learning (RL).
To address this exploration bottleneck, existing studies have mainly pursued two directions: reparameterization-based exploration and latent-explicit hybrid reasoning. One line of work applies reparameterization techniques~\citep{think-silently,soft-token-hard-truth,soft-grpo} to inject stochasticity in continuous representations, typically through Gaussian or Gumbel noise. 
Another line of research seeks to preserve the inherent diversity of discrete token distributions through hybrid architectures~\citep{hrpo,multiplex,swireasoning,hyrea}, which combine latent reasoning with discrete token generation via dense integration or mode switching. 
However, existing switching approaches often rely on rigid heuristic rules~\citep{swireasoning,thinkrouter} or supervised initialization~\citep{LiteReason,hyrea}. 
Consequently, there remains a critical gap for a lightweight, end-to-end RL framework capable of adaptive fine-grained mode switching.




In this paper, we propose TARPO (\textbf{T}oken-Wise Latent-Explicit Reasoning via \textbf{A}ction-\textbf{R}outing \textbf{P}olicy \textbf{O}ptimization), a pure RL framework that enables token-wise transitions between discrete and latent reasoning mode. 
Specifically, we introduce a lightweight action head router that observes the current hidden state to dynamically determine, at each step, whether the next reasoning unit should be constructed from a discrete token embedding or a continuous latent representation. 
Importantly, by formulating this routing decision as a choice within a discrete action space, our approach leverages the intrinsic stochasticity of routing policies to induce exploration over reasoning modes. 
Following the GRPO~\citep{grpo} paradigm, we jointly optimize the action head router and the LLM backbone end-to-end using a shared advantage signal. This unified objective enables the model to autonomously learn adaptive reasoning strategies without relying on heuristic rules or supervised initialization. 
Extensive experiments across Qwen2.5 backbones (from 1.5B to 7B) demonstrate that TARPO outperforms existing latent reasoning RL baselines while cross-architecture evaluation on Llama-3.1-8B confirms its generalization on three mathematical benchmarks. Out-of-distribution results on Qwen2.5-3B further show improved generalization and token efficiency, and our analysis reveals adaptive switching between discrete and latent reasoning.



The main contributions of this work are summarized as follows: 
\begin{itemize}[itemsep=-1.5pt, topsep=0pt]

    \item We propose TARPO, a pure reinforcement learning framework for token-wise latent-explicit reasoning, which adaptively switches between discrete token generation and continuous latent reasoning.
    
    
    

    \item We formulate reasoning-mode selection as a learnable action-routing policy and introduce a joint optimization objective that end-to-end trains the router and the LLM backbone.
    
    

    
    
    \item Extensive experiments and analyses demonstrate that TARPO consistently outperforms existing explicit and latent reasoning RL baselines, while learning adaptive switching behaviors across diverse benchmarks.

\end{itemize}

\section{Related Work}


\subsection{From Discrete CoT to Latent Reasoning}
Standard Chain-of-Thought (CoT)~\citep{think-step-by-step,cot} elicits reasoning through discrete tokens, which imposes an information bottleneck that limits representational capacity~\citep{coconut}. 
To overcome this, recent work shifts reasoning into the latent space~\citep{survey1,survey2,hidden-thinking,token-assorted}. 
Within this paradigm, continuous latent reasoning methods condition reasoning steps directly on transformer hidden states~\citep{coconut,system1.5,recurrent-depth} or probability-weighted token embedding mixtures~\citep{soft-thinking}. 
For instance, COCONUT~\citep{coconut} directly feeds hidden states into subsequent steps to enable parallel path exploration, while System-1.5~\citep{system1.5} adaptively reuses these representations via dynamic shortcuts. 
However, directly utilizing raw hidden states often causes representation manifold mismatch~\citep{soft-thinking,hrpo} and catastrophic forgetting~\citep{softcot}. Consequently, most recent architectures adopt token embedding mixtures. Nevertheless, these continuous representations remain inherently deterministic, limiting the intrinsic exploration capability required for reinforcement learning.



\subsection{Exploration in Latent Reasoning RL}


\paragraph{Reparameterization-Based Exploration.}
This line of work applies reparameterization techniques to introduce stochasticity into continuous representations, enabling policy exploration and gradient-based optimization. 
For instance, some recent methods inject Gaussian noise into compressed latent variables~\citep{think-silently,Reinforced-latent-reasoning} or continuous token embeddings~\citep{soft-token-hard-truth}. Meanwhile, other approaches~\citep{llm-as-a-single-threaded,soft-grpo,lepo,latent-grpo} apply the Gumbel-Softmax trick to derive probabilistic soft-token distributions. 


\begin{figure*}[t]
  \centering
  \begin{subfigure}[b]{0.50\linewidth}
    \centering
    \includegraphics[width=\linewidth]{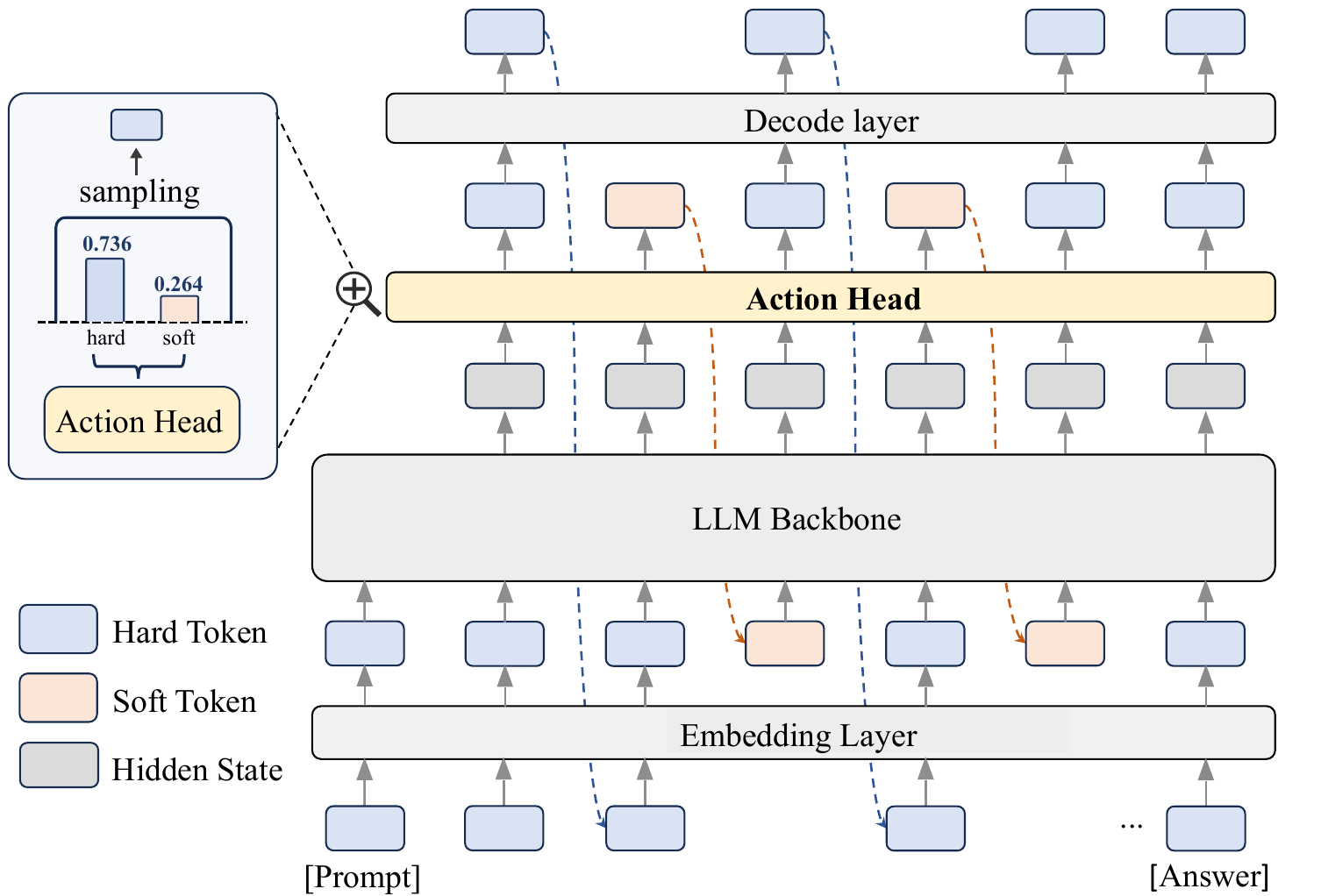}
    \caption{Token-Wise Action Routing}
    \label{fig:method-a}
  \end{subfigure}
  \hfill
  \begin{subfigure}[b]{0.49\linewidth}
    \centering
    \includegraphics[width=\linewidth]{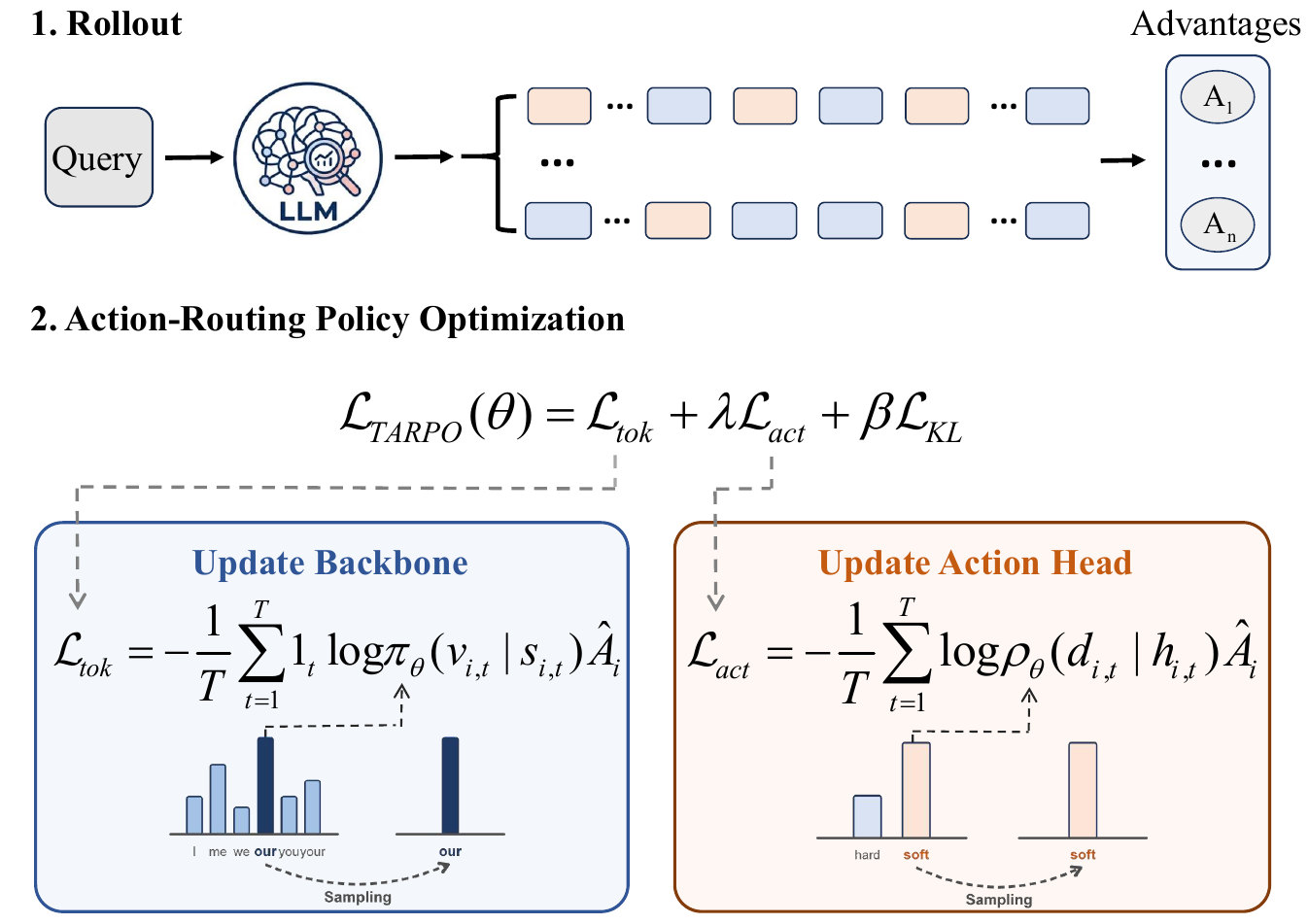}
    \caption{Action-Routing Policy Optimization}
    \label{fig:method-b}
  \end{subfigure}
  \caption{Overview of the TARPO framework. (a) During reasoning, a lightweight Action Head receives the current hidden state and routes the next step to either discrete token generation (\texttt{hard}) or continuous latent reasoning (\texttt{soft}). (b) The framework is trained end-to-end with a shared group-relative advantage signal, which jointly updates the LLM backbone and the action head from sampled hybrid rollouts.}
  \label{fig:method}
\end{figure*}



\paragraph{Latent-Explicit Hybrid Reasoning.}
As an alternative to reparameterization, this line of work seeks to fuse explicit discrete tokens with continuous latent representations, preserving the intrinsic stochasticity of discrete token sampling. 
One paradigm constructs a dense representation within a continuous space at each decoding step. 
For instance, HRPO~\citep{hrpo} blends hidden state representations with token embeddings via a learnable gate, while Multiplex Thinking~\citep{multiplex} aggregates multiple independently sampled tokens into a single continuous token. Distinct from these dense integration approaches, another branch explores selective reasoning mode switching along the discrete trajectory. However, existing frameworks typically rely on rigid heuristics such as token entropy~\citep{swireasoning,thinkrouter,selar,DyLaR} or heavily supervised initialization~\citep{think-at-hard,LiteReason,hyrea}, which limit autonomous exploration. 




Motivated by these limitations, TARPO introduces a token-wise action-routing mechanism that learns adaptive mode switching entirely through reinforcement learning. Unlike prior hybrid methods that rely on fixed heuristics or supervised initialization to determine when to switch between discrete and latent reasoning, TARPO formulates reasoning-mode selection as a learnable discrete routing policy over a binary action space. This design preserves the intrinsic stochasticity of discrete token sampling while enabling fine-grained transitions at each step, and allows for end-to-end optimization of the routing head and LLM backbone within a unified RL objective.


\section{Methodology}
\label{sec:method}



\subsection{Problem Formulation}
\label{subsec:problem_formulation}

We formulate token-wise latent-explicit reasoning as a sequential decision process.
At step $t$, the state $s_t$ corresponds to the current context history
$s_t = (x, u_1, u_2, \ldots, u_{t-1})$, where $x$ is the initial user prompt and
each reasoning unit $u_i$ is either a discrete token embedding or a continuous latent
representation.

To enable adaptive transitions between explicit and latent reasoning, we define a binary routing decision space $\mathcal{D} = \{\texttt{hard}, \texttt{soft}\}$ and a joint action space:
\begin{equation}
\mathcal{A} = \{\texttt{soft}\} \cup \left(\{\texttt{hard}\} \times \mathcal{V}\right),
\end{equation}
where $\mathcal{V}$ is the vocabulary, $\texttt{hard}$ denotes explicit token reasoning, and $\texttt{soft}$ denotes latent reasoning in continuous space. At each step, the routing policy $\rho_\theta$ over $\mathcal{D}$ first selects the reasoning mode, and the language policy $\pi_\theta$ over $\mathcal{V}$ is applied only when $\texttt{hard}$ is chosen. Together, they define the full action distribution over $\mathcal{A}$, determining how the next reasoning unit $u_t$ is constructed and appended to the context history.

\subsection{Token-Wise Routing for Latent and Explicit Reasoning}


As shown in Figure~\ref{fig:method-a}, we introduce a lightweight action head on top of the LLM backbone to route the reasoning mode at each token step. Given the final hidden state, the action head predicts a binary routing action that determines whether the next reasoning unit is constructed from a discrete token embedding or a continuous latent vector. 


\paragraph{The Routing Policy.}
Let $h_t \in \mathbb{R}^{d}$ denote the hidden state produced by the final transformer layer at step $t$. We instantiate the router as a lightweight linear projection head that directly maps the hidden state into a categorical distribution over the reasoning actions. The routing policy $\rho_{\theta}$ is parameterized as:
\begin{equation}
    \rho_{\theta}(\cdot | h_t) = \text{Softmax}(\mathbf{W}_r h_t + \mathbf{b}_r)
\end{equation}
where $\mathbf{W}_r \in \mathbb{R}^{2 \times d}$ and $\mathbf{b}_r \in \mathbb{R}^{2}$ are learnable parameters. During training, we sample the routing decision $d_t \sim \rho_{\theta}(\cdot \mid h_t)$ to encourage structural exploration. At inference time, the router supports both stochastic and deterministic routing strategies. We initialize the action-head bias $\mathbf{b}_r$ to $b_0$ to mildly favor hard routing, consistent with the model's preference, which we further discuss in Section~\ref{sec:bias_kl} and Appendix~\ref{app:bias_kl}. 


\paragraph{Hybrid State Transition.}
The routing decision determines how the next unit $u_t$ is constructed. Specifically, if $d_t$ selects explicit reasoning, the model appends the discrete token embedding $u_t = \mathbf{E}(v_t)$; otherwise, it constructs a latent unit by a sparse weighted sum over the top-$k$ tokens:
\begin{equation}
    u_t =
    \begin{cases}
        \mathbf{E}(v_t), & \text{if } d_t = \texttt{hard}, \\
        \sum_{i \in \mathcal{K}_t} w_t^{(i)} \mathbf{E}(v^{(i)}), & \text{if } d_t = \texttt{soft},
    \end{cases}
\end{equation}
where $\mathcal{K}_t$ denotes the top-$k$ token set and $w_t^{(i)}$ is normalized by the softmax over the corresponding logits. This hybrid routing is applied only during the thinking phase, while the final answer after the answer tag is generated as standard readable discrete tokens. 


\subsection{Action-Routing Policy Optimization}

We optimize TARPO with an online group-relative policy optimization objective over hybrid reasoning trajectories. For each prompt, we sample a group of trajectories and assign each trajectory a scalar outcome reward based on the final generated answer. We normalize the rewards within each group and use the same group-relative advantage signal to optimize both the LLM backbone and the action head router. 

Let $r_i$ denote the reward of the $i$-th sampled trajectory in a group. The shared group-relative advantage is computed as:
\begin{equation}
\hat A_i
=
\frac{r_i-\mu_r}{\sigma_r+\epsilon},
\end{equation}
where $\mu_r$ and $\sigma_r$ denote the mean and standard deviation of the sampled rewards within the group, respectively.


Given the generated tokens $y_{i,1:T}$ and routing decisions $d_{i,1:T}$, we jointly optimize the backbone policy and the routing policy.
Here, $\mathcal{L}_{\text{tok}}^{(i)}$ denotes the token-generation objective of the LLM backbone for the $i$-th trajectory, $\mathcal{L}_{\text{act}}^{(i)}$ denotes the action-routing objective of the router, and $\mathcal{L}_{\text{KL}}^{(i)}$ denotes the KL regularization term.
The overall objective averages over the sampled group:
\begin{equation}
    \mathcal{L}_{\text{TARPO}}
    = \frac{1}{G} \sum_{i=1}^{G}
      \Bigl[
        \mathcal{L}_{\text{tok}}^{(i)}
        + \lambda\, \mathcal{L}_{\text{act}}^{(i)}
        + \beta\, \mathcal{L}_{\text{KL}}^{(i)}
      \Bigr],
\end{equation}
where $\lambda$ controls the contribution of the routing objective and $\beta$ is the base coefficient for the KL penalty. 
Let $\delta_t = \rho_\theta(\texttt{Hard}\mid h_t)$ denote the probability of taking the hard path at step $t$:
\begin{multline}
\label{eq:kl_loss}
\mathcal{L}_{\text{KL}}^{(i)} = \frac{1}{T}\sum_{t=1}^{T} \Big[
  \delta_t\, D_{\text{KL}}\!\left(\pi_\theta(\cdot\mid s_t) \,\|\, \pi_{\text{ref}}(\cdot\mid s_t)\right) \\
  + \alpha\, D_{\text{KL}}\!\left(\rho_\theta(\cdot\mid h_t) \,\|\, \rho_{\text{ref}}(\cdot\mid h_t)\right)
\Big],
\end{multline}
where $\alpha$ controls the relative strength of the KL regularization applied to the action head. Setting $\alpha=0$ removes the action KL entirely, which we analyze in Section~\ref{sec:bias_kl} and Appendix~\ref{app:bias_kl}.

The token-generation and routing objectives are optimized using the shared advantage signal:
\begin{align}
    \mathcal{L}_{\text{tok}}^{(i)}
    &= -\frac{1}{T_i} \sum_{t=1}^{T_i}
       \mathbf{1}_t\,
       \log \pi_{\theta}(v_{i,t} \mid s_{i,t})\,
       \hat{A}_i,
    \label{eq:loss_tok} \\
    \mathcal{L}_{\text{act}}^{(i)}
    &= -\frac{1}{T_i} \sum_{t=1}^{T_i}
       \log \rho_{\theta}(d_{i,t} \mid h_{i,t})\,
       \hat{A}_i,
    \label{eq:loss_act}
\end{align}
where $\mathbf{1}_t=\mathbf{1}[d_{i,t}=\texttt{Hard}]$ indicates that the router selects the hard pattern at step $t$.


This objective jointly updates the LLM backbone and the action head router, encouraging successful reasoning trajectories while learning adaptive latent-explicit routing patterns.

\section{Experiment}
\label{sec:experiment}

\begin{table*}[!t]
\centering
\small
\resizebox{\textwidth}{!}{ 
\begin{tabular}{l cccccccccccc} 
\toprule
\multirow{2}{*}{Method} & \multicolumn{2}{c}{GSM8K} & \multicolumn{2}{c}{MATH} & \multicolumn{2}{c}{MATH500} & \multicolumn{2}{c}{AMC23} & \multicolumn{2}{c}{Olympiad} & \multicolumn{2}{c}{Average} \\
\cmidrule(lr){2-3} \cmidrule(lr){4-5} \cmidrule(lr){6-7} \cmidrule(lr){8-9} \cmidrule(lr){10-11} \cmidrule(lr){12-13}
& P@1 & P@32 & P@1 & P@32 & P@1 & P@32 & P@1 & P@32 & P@1 & P@32 & P@1 & P@32 \\
\midrule
\midrule
\multicolumn{13}{c}{Qwen2.5-1.5B-Instruct} \\ 
\midrule
CoT & 60.89 & 95.53 & 31.66 & 78.90 & 31.76 & 78.60 & 18.83 & 70.00 & 14.13 & 49.04 & 31.45 & 74.41 \\
Pure Latent & 60.53 & 65.50 & 25.29 & 37.20 & 26.51 & 38.60 & 15.86 & 20.00 & 12.21 & 17.33 & 28.08 & 35.73 \\
Entropy-Routed & 60.66 & 93.10 & 25.29 & 72.04 & 27.10 & 71.80 & 13.98 & 55.00 & 12.65 & 46.67 & 27.94 & 67.72 \\
\midrule
GRPO & 69.04 & 96.66 & 47.82 & 81.22 & 48.56 & 81.20 & \textbf{26.80} & 77.50 & 16.42 & \textbf{50.37} & 41.73 & 77.39 \\
HRPO & 69.71 & 96.66 & 48.99 & \textbf{82.34} & \textbf{49.80} & 82.60 & 25.78 & 72.50 & \textbf{16.51} & 48.44 & 42.16 & 76.51 \\
\textbf{TARPO (Ours)} & \textbf{70.76} & \textbf{96.97} & \textbf{49.71} & 82.28 & 49.71 & \textbf{83.40} & 25.47 & \textbf{80.00} & 16.13 & 48.89 & \textbf{42.36} & \textbf{78.31} \\
\midrule
\midrule
\multicolumn{13}{c}{Qwen2.5-3B-Instruct} \\
\midrule
CoT & 73.50 & 96.97 & 55.64 & 85.00 & 56.02 & 84.40 & 36.48 & 82.50 & 22.83 & 54.07 & 48.89 & 80.59 \\
Pure Latent & 73.84 & 78.92 & 55.45 & 59.70 & 56.70 & 60.20 & \textbf{43.20} & 45.00 & \textbf{24.85} & 28.00 & 50.81 & 54.36 \\
Entropy-Routed & 73.91 & 95.53 & 54.75 & 83.36 & 55.57 & 82.20 & 37.58 & 80.00 & 22.99 & 53.04 & 48.96 & 78.83 \\
\midrule
GRPO & 82.55 & \textbf{97.57} & 59.86 & 84.72 & \textbf{60.85} & 84.40 & 38.67 & 87.50 & 23.58 & 54.07 & 53.10 & 81.65 \\
HRPO & 82.75 & 97.42 & \textbf{59.97} & 85.06 & 60.64 & \textbf{85.00} & 38.83 & 85.00 & 23.80 & 55.26 & 53.20 & 81.55 \\
\textbf{TARPO (Ours)} & \textbf{83.23} & 96.97 & 59.75 & \textbf{85.20} & 60.57 & 84.80 & 40.23 & \textbf{90.00} & 24.27 & \textbf{57.04} & \textbf{53.61} & \textbf{82.80} \\
\midrule
\midrule
\multicolumn{13}{c}{Qwen2.5-7B-Instruct} \\
\midrule
CoT & 87.97 & \textbf{97.73} & 61.16 & 87.30 & 61.51 & 86.00 & 44.22 & 90.00 & 25.73 & 56.15 & 56.12 & 83.43 \\
Pure Latent & 88.40 & 90.90 & 56.78 & 68.66 & 59.50 & 70.80 & 48.52 & 60.00 & 27.55 & 34.52 & 56.15 & 64.98 \\
Entropy-Routed & 88.07 & 97.04 & 54.02 & 82.96 & 53.91 & 81.39 & 41.56 & 82.50 & 26.58 & \textbf{56.74} & 52.83 & 80.13 \\
\midrule
GRPO & 89.87 & 97.42 & 70.05 & 88.02 & 69.86 & \textbf{87.80} & 51.72 & 87.50 & \textbf{30.96} & 53.78 & 62.49 & 82.90 \\
HRPO & \textbf{90.13} & 97.42 & \textbf{70.25} & 88.20 & 69.78 & 87.60 & 51.17 & 85.00 & 30.75 & 53.19 & 62.42 & 82.28 \\
\textbf{TARPO (Ours)} & 89.94 & 97.27 & 70.22 & \textbf{88.22} & \textbf{70.26} & \textbf{87.80} & \textbf{53.52} & \textbf{95.00} & 30.69 & 54.22 & \textbf{62.92} & \textbf{84.50} \\
\bottomrule
\end{tabular}
}
\caption{Main results on the reasoning benchmarks. We compare our method with discrete baselines and prior latent reasoning methods across different base models. The best results overall are highlighted in bold. P@$k$ denotes Pass@$k$ accuracy.}
\label{tab:main_results}
\end{table*}

\begin{table}[!t]
\centering
\small
\setlength{\tabcolsep}{4pt} 
\begin{tabular}{l cc cc cc}
\toprule
\multirow{2}{*}{Method} & \multicolumn{2}{c}{GSM8K} & \multicolumn{2}{c}{MATH500} & \multicolumn{2}{c}{AMC23} \\
\cmidrule(lr){2-3} \cmidrule(lr){4-5} \cmidrule(lr){6-7}
& P@1 & P@32 & P@1 & P@32 & P@1 & P@32 \\
\midrule
\midrule
CoT & 77.78 & \textbf{96.82} & 35.48 & 77.20 & 16.17 & \textbf{77.50} \\
GRPO & 82.98 & \textbf{96.82} & 45.77 & 81.80 & 21.56 & 70.00 \\
\textbf{TARPO} & \textbf{84.25} & 95.91 & \textbf{47.65} & \textbf{83.20} & \textbf{22.97} & 62.50 \\
\bottomrule
\end{tabular}
\caption{Generalization performance on the Llama-3.1-8B-Instruct architecture. Models are trained on the MATH dataset.
}
\label{tab:Llama_generalization}
\end{table}

\subsection{Experimental Settings}

\paragraph{Training \& Testing Settings.}
We conduct experiments primarily on three backbones including Qwen2.5-1.5B-Instruct, Qwen2.5-3B-Instruct, and Qwen2.5-7B-Instruct. 
We further provide supplementary experiments on Llama-3.1-8B-Instruct to evaluate the cross-architecture generalization of TARPO. 
For in-domain evaluation, models trained on GSM8K are evaluated on GSM8K, while models trained on MATH are evaluated on MATH, MATH500, AMC23, and OlympiadBench. 
For out-of-distribution (OOD) evaluation, we train on the DAPO-MATH-17k~\citep{dapo-math} dataset and evaluate on the scientific reasoning benchmarks GPQA-Diamond~\citep{gpqa} and ARC-C~\citep{arcc}, as well as the code generation benchmark HumanEval~\citep{humaneval}. 
We follow the overall training framework and implementation setup of HRPO~\citep{hrpo}. Detailed training configurations and generation settings are provided in Appendix~\ref{app:setting}.

\paragraph{Baselines.}
We compare TARPO with both representative training-free and RL-based reasoning paradigms. For training-free baselines, we include: (1) \textbf{CoT}, standard Chain-of-Thought prompting; (2) \textbf{Pure Latent}, a basic latent reasoning mechanism in which all reasoning steps are performed with soft tokens; and (3) \textbf{Entropy-Routed}, a hybrid latent reasoning strategy that uses token entropy heuristics to trigger mode switching, inspired by prior entropy-guided routing methods~\citep{swireasoning}. Specifically, at each token step, if the token entropy exceeds a predefined threshold (set to 0.5 in our implementation), the next step is constructed using a soft token. For RL-based baselines, we compare against: (4) \textbf{GRPO}~\citep{grpo} as a representative discrete RL baseline; and (5) \textbf{HRPO}~\citep{hrpo} as a representative hybrid latent RL baseline, where a gating mechanism fuses discrete token embeddings and continuous representations at each step. We additionally include partial comparisons with reparameterization-based latent RL methods, such as \textbf{Soft Tokens}~\citep{soft-token-hard-truth}.


\begin{table}[!t]
\centering
\small
\setlength{\tabcolsep}{2pt} 
\begin{tabular}{l ccc ccc}
\toprule
\multirow{2}{*}{Method} & \multicolumn{3}{c}{GSM8K} & \multicolumn{3}{c}{MATH500} \\
\cmidrule(lr){2-4} \cmidrule(lr){5-7}
& Greedy & P@1 & P@32 & Greedy & P@1 & P@32 \\
\midrule
\midrule
CoT & 75.82 & 73.50 & 96.97 & 56.60 & 56.02 & 84.40 \\
GRPO & 83.70 & 82.53 & \textbf{97.73} & 61.60 & \textbf{60.85} & 84.40 \\
GRPO* & \textbf{84.00} & \textbf{83.00} & 97.20 & 59.00 & 57.10 & 83.60 \\
\midrule
Soft Tokens* & 82.90 & 81.90 & 95.40 & 54.70 & 54.60 & 80.30 \\
\textbf{TARPO} & 83.24 & 82.44 & 97.12 & \textbf{62.00} & 60.57 & \textbf{84.80} \\
\bottomrule
\end{tabular}
\caption{Comparison with the reparameterization-based method on Qwen2.5-3B-Instruct. We train on the same MATH dataset. Results marked with * are taken from \citet{soft-token-hard-truth}.}
\label{tab:reparam_comparison}
\end{table}

\begin{table*}[t]
\centering
\small 
\resizebox{\textwidth}{!}{ 
\begin{tabular}{l ccc ccc ccc ccc} 
\toprule
\multirow{2}{*}{Method} & \multicolumn{3}{c}{GPQA-Diamond} & \multicolumn{3}{c}{ARC-C} & \multicolumn{3}{c}{HumanEval} & \multicolumn{3}{c}{Average} \\
\cmidrule(lr){2-4} \cmidrule(lr){5-7} \cmidrule(lr){8-10} \cmidrule(lr){11-13}
& P@1 & P@32 & \#Tok & P@1 & P@32 & \#Tok & P@1 & P@32 & \#Tok & P@1 & P@32 & \#Tok \\
\midrule
\midrule
CoT            & 26.83 & \textbf{92.93} & 410.6 & 74.05 & 98.21 & 106.2 & 59.68 & 90.24 & 271.4 & 53.52 & \textbf{93.79} & 262.7 \\
Pure Latent    & 27.26 & 37.37 & 405.6 & 70.57 & 78.16 & 97.1  & 60.19 & 82.32 & 261.4 & 52.67 & 65.95 & 254.7 \\
Entropy-Routed & 28.30 & 90.91 & 373.2 & 71.09 & 96.25 & 95.4  & 58.31 & 90.24 & 260.5 & 52.57 & 92.47 & 243.0 \\
\midrule
GRPO           & 28.03 & 83.33 & 608.5 & 74.78 & 98.12 & 272.4 & 58.86 & \textbf{92.07} & 355.6 & 53.89 & 91.17 & 412.2 \\
HRPO           & 27.51 & 88.38 & 603.3 & \textbf{75.06} & 97.44 & 277.8 & 59.51 & 90.85 & 340.5 & 54.03 & 92.22 & 407.2 \\
\textbf{TARPO} & \textbf{28.41} & 90.91 & \textbf{568.3} & 74.01 & \textbf{98.89} & \textbf{189.4} & \textbf{63.62} & 88.41 & \textbf{256.1} & \textbf{55.35} & 92.74 & \textbf{337.9} \\
\bottomrule
\end{tabular} 
}
\caption{ Out-of-distribution (OOD) generalization evaluation results of Qwen2.5-3B-Instruct. We report pass@1, pass@32, and generated tokens on GPQA-Diamond, ARC-C, and HumanEval datasets.}
\label{tab:ood_results}
\end{table*}

\begin{table*}[!t]
\centering
\small
\setlength{\tabcolsep}{4pt}
\begin{tabular}{l cccccccccccc}
\toprule
\multirow{2}{*}{Method} & \multicolumn{2}{c}{GSM8K} & \multicolumn{2}{c}{MATH} & \multicolumn{2}{c}{MATH500} & \multicolumn{2}{c}{AMC23} & \multicolumn{2}{c}{Olympiad} & \multicolumn{2}{c}{Average} \\
\cmidrule(lr){2-3} \cmidrule(lr){4-5} \cmidrule(lr){6-7} \cmidrule(lr){8-9} \cmidrule(lr){10-11} \cmidrule(lr){12-13}
& P@1 & P@32 & P@1 & P@32 & P@1 & P@32 & P@1 & P@32 & P@1 & P@32 & P@1 & P@32 \\
\midrule
\midrule
\multicolumn{13}{c}{Qwen2.5-1.5B-Instruct} \\
\midrule
GRPO & 69.04 & 96.66 & 47.82 & 81.22 & 48.56 & 81.20 & \textbf{26.80} & 77.50 & 16.42 & \textbf{50.37} & 41.73 & 77.39 \\
$\quad$ \textit{w/ Pure Latent} & \textbf{71.21} & 71.57 & 48.16 & 48.98 & 48.97 & 49.20 & 17.50 & 17.50 & \textbf{16.56} & 17.63 & 40.48 & 40.98 \\
$\quad$ \textit{w/ Entropy Routing} & 70.23 & 94.24 & 47.21 & 78.54 & 48.04 & 79.20 & 24.53 & 70.00 & 16.13 & 46.37 & 41.23 & 73.67 \\
TARPO & 70.76 & \textbf{96.97} & \textbf{49.71} & \textbf{82.28} & \textbf{49.71} & \textbf{83.40} & 25.47 & \textbf{80.00} & 16.13 & 48.89 & \textbf{42.36} & \textbf{78.31} \\
\midrule
\midrule
\multicolumn{13}{c}{Qwen2.5-3B-Instruct} \\
\midrule
GRPO & 82.55 & \textbf{97.57} & \textbf{59.86} & 84.72 & \textbf{60.85} & 84.40 & 38.67 & 87.50 & 23.58 & 54.07 & 53.10 & 81.65 \\
$\quad$ \textit{w/ Pure Latent} & 82.64 & 83.40 & 59.44 & 60.48 & 60.65 & 61.60 & 39.69 & 40.00 & 23.66 & 24.89 & 53.22 & 54.07 \\
$\quad$ \textit{w/ Entropy Routing} & 82.94 & 95.98 & 59.53 & 83.72 & 60.64 & 83.80 & 38.67 & 77.50 & 24.25 & 52.44 & 53.21 & 78.69 \\
TARPO & \textbf{83.23} & 96.97 & 59.75 & \textbf{85.20} & 60.57 & \textbf{84.80} & \textbf{40.23} & \textbf{90.00} & \textbf{24.27} & \textbf{57.04} & \textbf{53.61} & \textbf{82.80} \\
\bottomrule
\end{tabular}
\caption{Ablation study on the reasoning benchmarks. We compare TARPO against GRPO and two GRPO-based ablated variants with different latent inference strategies: \textit{w/ Pure Latent}, which performs all reasoning steps in latent space, and \textit{w/ Entropy Routing}, which switches to latent space based on a fixed entropy threshold.}
\label{tab:ablation_results}
\end{table*}

\paragraph{Metrics.}
We report pass@1 and pass@32 as the primary evaluation metrics, where pass@1 is averaged over 32 sampled generations. For OOD benchmarks, we additionally report the average number of generated tokens, accounting for both discrete and latent tokens in TARPO. Additional detailed statistics are provided in Appendix~\ref{app:exp_stat}. 

\subsection{Results}

\paragraph{In-Domain Performance.}
We report the in-domain results in Table~\ref{tab:main_results}. Overall, TARPO surpasses both training-free baselines and RL-based methods across the three Qwen2.5 backbones. Averaged over the three scales, TARPO improves GRPO by 0.52\% and 1.22\% in Pass@1 and Pass@32, respectively, and outperforms HRPO by 0.37\% and 1.76\%. These results confirm that incorporating token-wise action routing provides a more effective and stable optimization landscape. 

\paragraph{OOD Generalization.}
We evaluate the Qwen2.5-3B-Instruct model on GPQA-Diamond, ARC-C, and HumanEval for out-of-distribution generalization. As shown in Table~\ref{tab:ood_results}, TARPO achieves the highest average Pass@1 (55.35\%), with a notable 4.76\% improvement over GRPO on HumanEval. Furthermore, TARPO significantly improves token efficiency, reducing the average generated tokens to 337.9 compared to over 400 for baseline RL methods, thereby maintaining concise reasoning trajectories.

\paragraph{Cross-Architecture Generalization.}
\label{sec:cross-architecture-results}
We further evaluate TARPO on the Llama-3.1-8B-Instruct backbone to assess its cross-architecture robustness. As shown in Table~\ref{tab:Llama_generalization}, TARPO achieves the best Pass@1 across all three benchmarks, outperforming GRPO by 1.27\%, 1.88\%, and 1.41\% on GSM8K, MATH500, and AMC23 respectively. TARPO also attains the highest Pass@32 on MATH500, indicating strong generalization across backbones.

\paragraph{Comparison with Reparameterization-Based Methods.} We additionally conduct a partial comparison with Soft Tokens~\citep{soft-token-hard-truth}, a reparameterization-based latent reasoning RL baseline. As shown in Table~\ref{tab:reparam_comparison}, TARPO achieves strong overall performance, improving MATH500 Pass@32 from 80.30\% to 84.80\%.





\begin{figure*}[t]
  \centering
  
  \begin{subfigure}{\linewidth}
    \centering
    \includegraphics[width=0.328\linewidth]{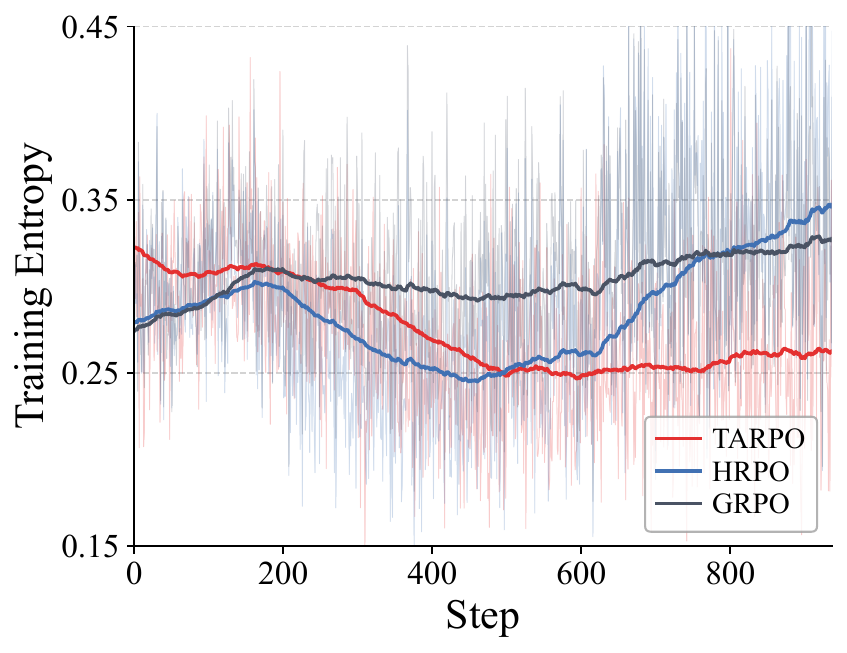}\hfill%
    \includegraphics[width=0.328\linewidth]{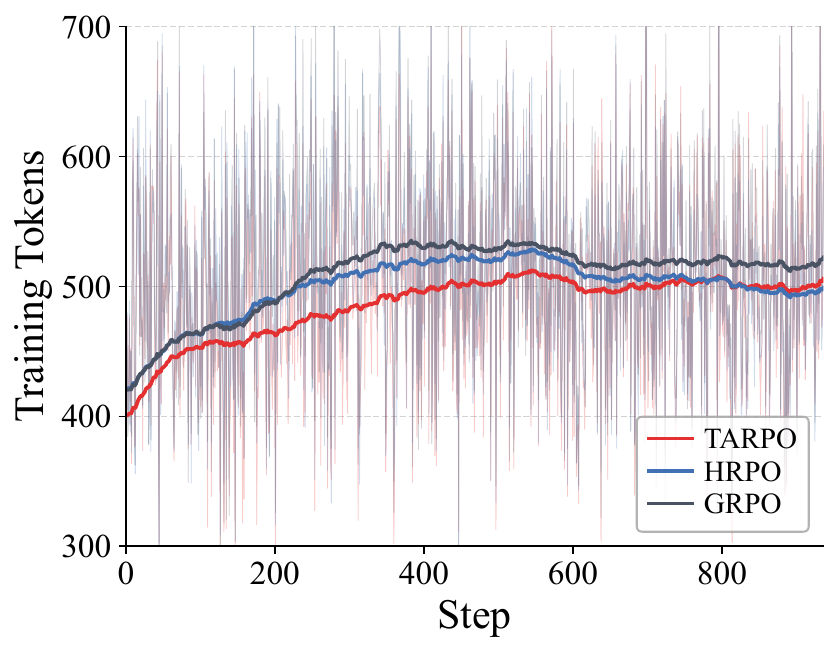}\hfill%
    \includegraphics[width=0.328\linewidth]{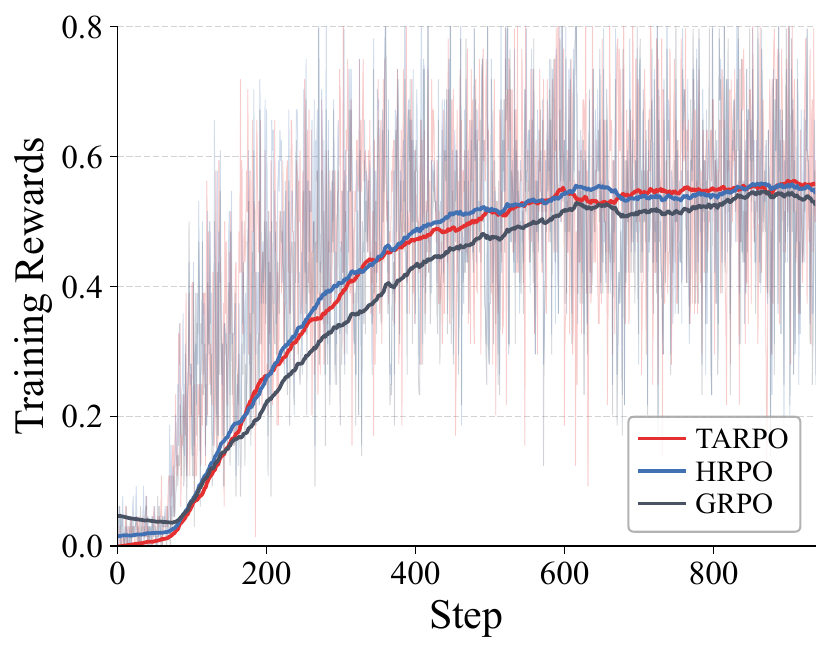}
    \caption{Training curves of Qwen2.5-1.5B-Instruct on MATH.}
    \label{fig:train_dyn_1.5b}
  \end{subfigure}

  \vspace{1em} 

  \begin{subfigure}{\linewidth}
    \centering
    \includegraphics[width=0.328\linewidth]{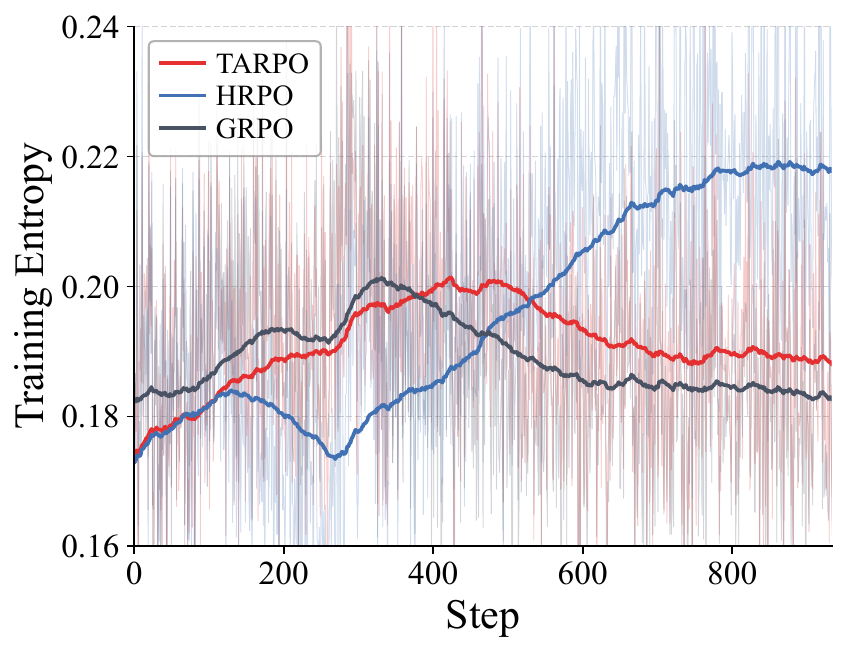}\hfill%
    \includegraphics[width=0.328\linewidth]{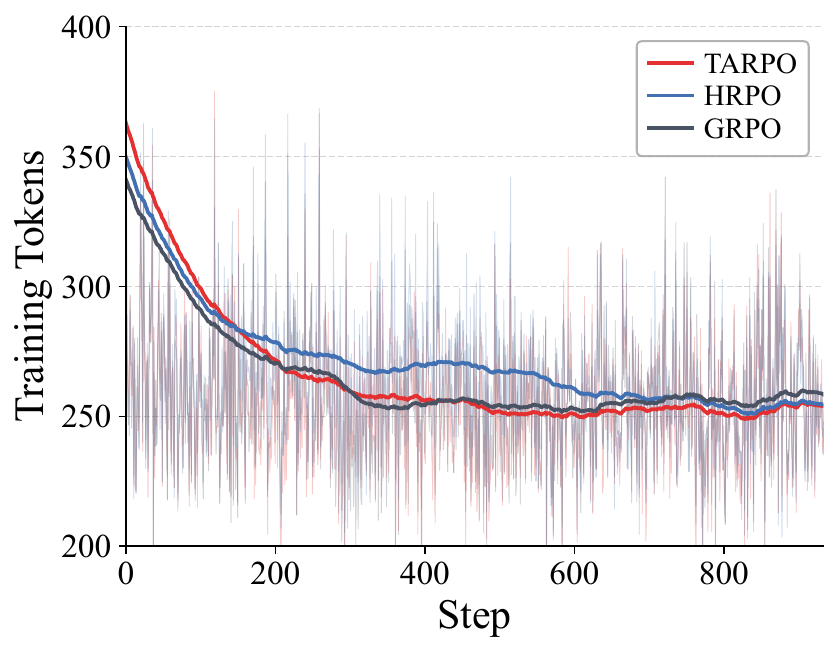}\hfill%
    \includegraphics[width=0.328\linewidth]{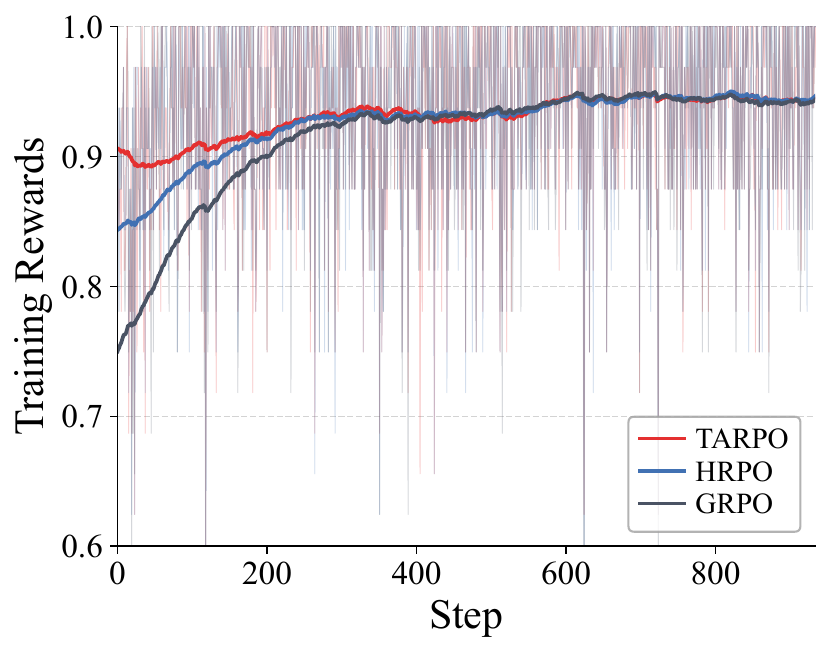}
    \caption{Training curves of Qwen2.5-7B-Instruct on GSM8K.}
    \label{fig:train_dyn_7b}
  \end{subfigure}

  \caption{
  Comparison of training curves between TARPO and baselines (GRPO, HRPO) across different model scales and datasets. 
  From left to right, the panels display Training Entropy, Training Tokens (counting both hard and soft tokens for TARPO), and Training Rewards.}
  \label{fig:training_dynamics}
\end{figure*}

\subsection{Ablation Study}
\label{sec:ablation}

To evaluate the learnable action-routing mechanism, we compare TARPO against GRPO and ablated variants: 
(1) \textbf{w/ Pure Latent}, performing all reasoning steps in the continuous latent space using soft tokens; and (2) \textbf{w/ Entropy Routing}, switching from discrete tokens to latent reasoning only when the current token entropy exceeds a predefined threshold (0.5 in our implementation). 


As shown in Table~\ref{tab:ablation_results}, TARPO consistently outperforms both ablated variants. For Qwen2.5-3B-Instruct, TARPO achieves the highest average Pass@1 of 53.61\% and Pass@32 of 82.80\%. Since soft-token reasoning is inherently deterministic, \textbf{w/ Pure Latent} can only rely on the final answer generation stage for stochastic exploration, which leads to a substantial drop in average Pass@32 to 54.07\%. In contrast, \textbf{w/ Entropy Routing} depends on a fixed heuristic trigger, making its switching behavior less adaptive. These results indicate that a learnable routing policy provides a more adaptive mechanism for balancing discrete and latent reasoning.

\subsection{Training Dynamics and Analysis}
\label{sec:training_dynamics}

\begin{figure*}[t]
  \centering
  \begin{subfigure}{0.328\linewidth}
    \includegraphics[width=\linewidth]{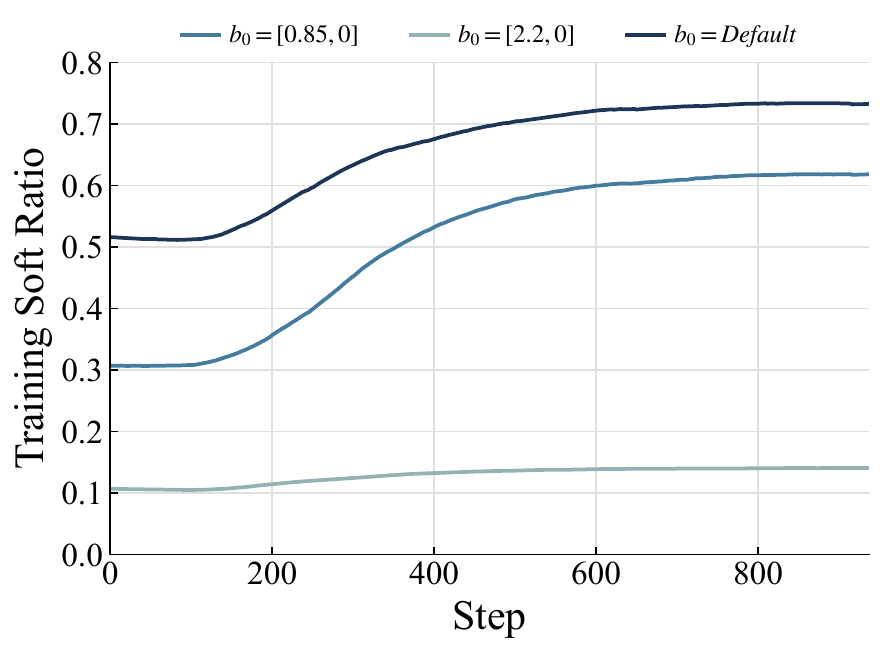}
    \caption{Impact of Initial Bias on Soft Ratio}
    \label{fig:bias_kl_a}
  \end{subfigure}
  \hfill%
  \begin{subfigure}{0.328\linewidth}
    \includegraphics[width=\linewidth]{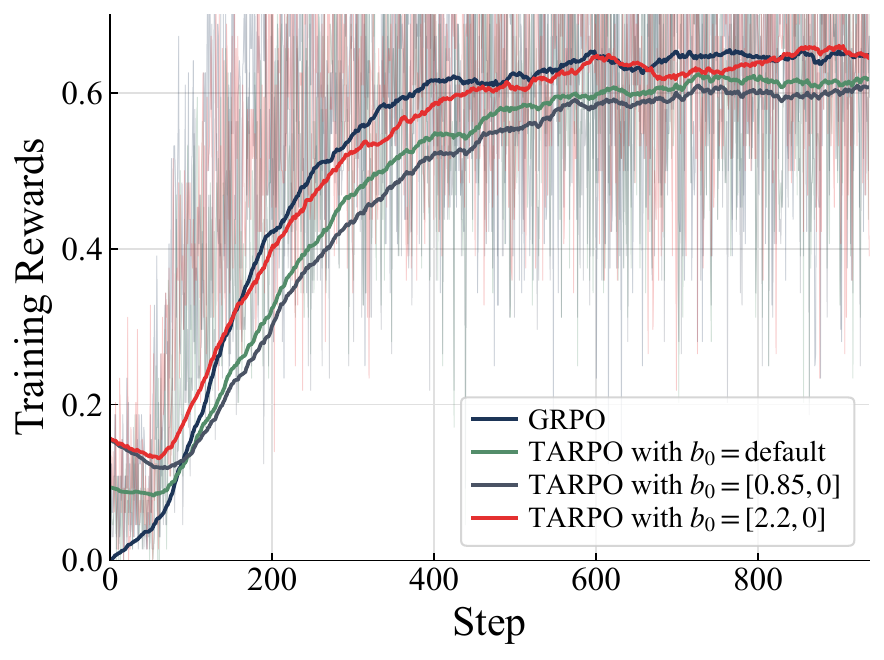}
    \caption{Impact of Initial Bias on Reward}
    \label{fig:bias_kl_b}
  \end{subfigure}
  \hfill%
  \begin{subfigure}{0.328\linewidth}
    \includegraphics[width=\linewidth]{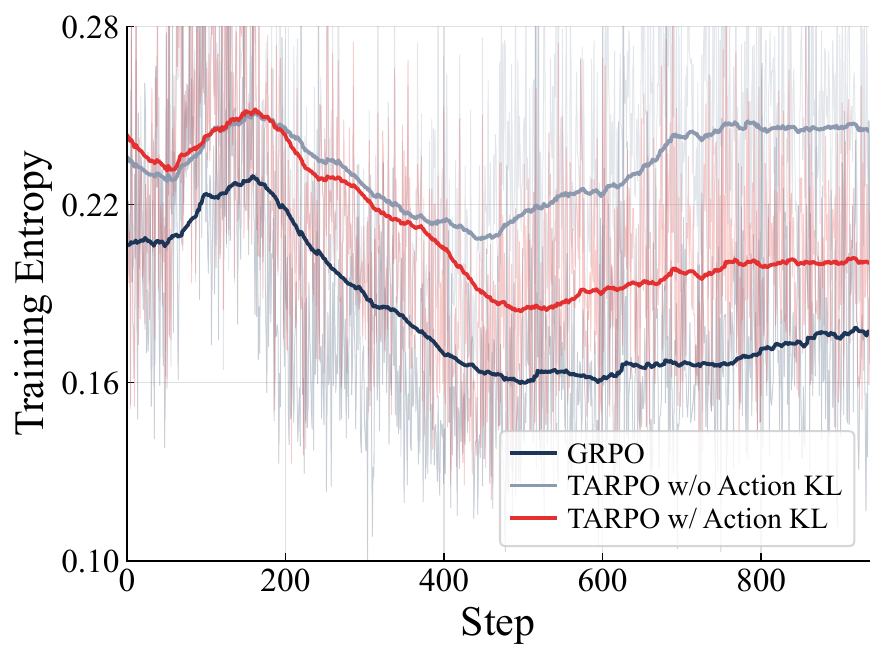}
    \caption{Effect of Action KL on Entropy}
    \label{fig:bias_kl_c}
  \end{subfigure}
 \caption{
 Analysis of TARPO action head bias initialization and KL (Qwen2.5-3B-Instruct on MATH).
 (a) and (b) show the impact of the initial bias ($b_{0}$) on the soft token ratio and training reward. (c) illustrates the effect of the action KL penalty on token entropy.}
  \label{fig:bias_kl}
\end{figure*}


\begin{figure*}[t]
  \centering
  \includegraphics[width=1.0\linewidth]{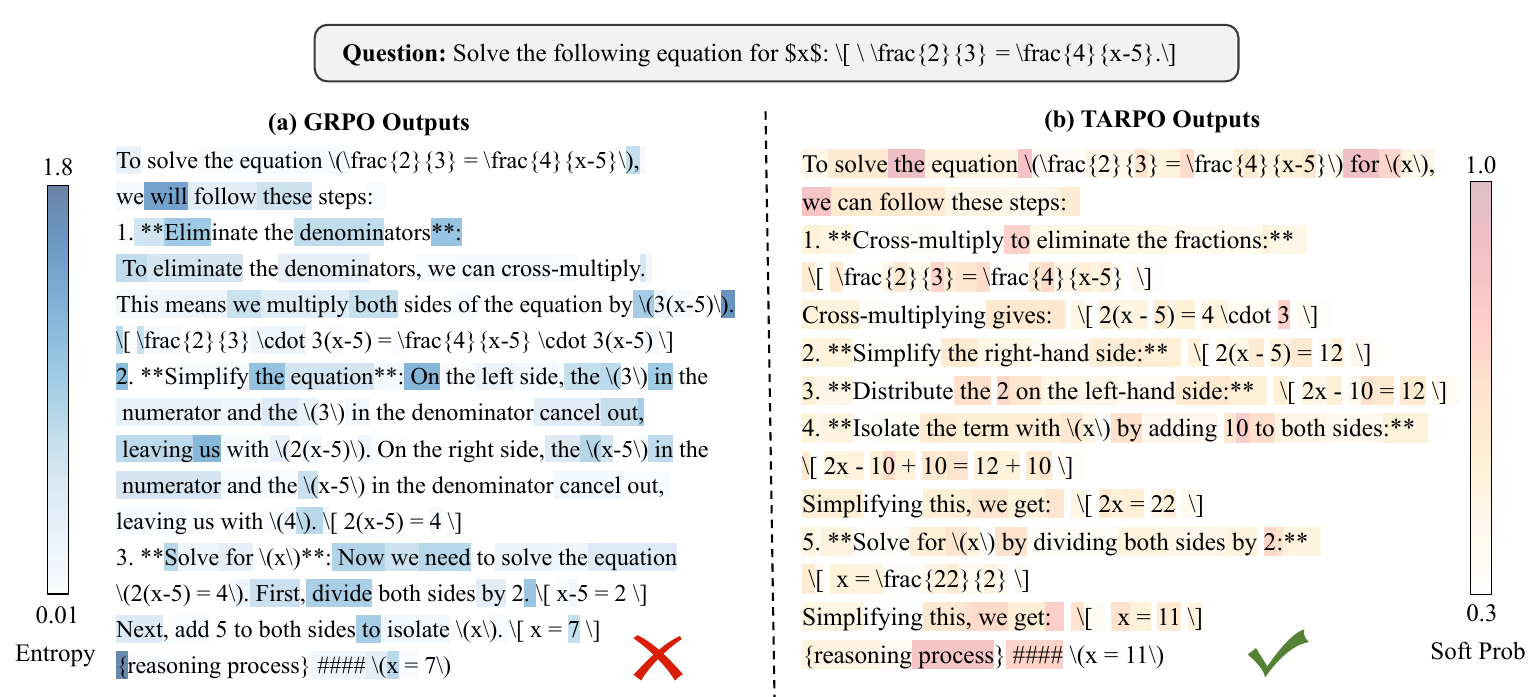}
  \caption{A case study on a MATH500 problem. (a) GRPO outputs with token entropy highlighted in blue. (b) TARPO outputs with soft probability highlighted in color.}
  \label{fig:case}
\end{figure*}

\paragraph{Entropy and Completion Length.}
As shown in Figure~\ref{fig:training_dynamics}, we track token entropy, completion length, and rewards during optimization to understand the training dynamics. On the 1.5B model trained on MATH in Figure~\ref{fig:train_dyn_1.5b}, TARPO consistently produces shorter completions while achieving rewards comparable to HRPO and slightly higher than GRPO, indicating better token efficiency at similar performance. On the 7B model trained on GSM8K in Figure~\ref{fig:train_dyn_7b}, HRPO exhibits an entropy rise in the later training stage, whereas TARPO largely follows the GRPO trajectory but maintains a moderately higher entropy level near convergence, suggesting a better balance between exploration and efficiency.


\paragraph{Action Head Bias and KL.}
\label{sec:bias_kl}
To further analyze the internal dynamics of the learnable routing mechanism, we investigate the impact of the action head initial bias ($b_0$) and the action KL penalty using Qwen2.5-3B-Instruct on the MATH dataset. Specifically, under the penalty-free setting (w/o Action KL), Figure~\ref{fig:bias_kl_a} and~\ref{fig:bias_kl_b} show that different initial bias settings result in different soft token ratios and reward dynamics during training, indicating that the router actively adapts its reasoning preference under RL optimization. 
Figure~\ref{fig:bias_kl_c} shows the effect of the action KL penalty under the initial bias setting of $b_0 = [2.2, 0]$, comparing the default ($\alpha=1$, w/ Action KL) and penalty-free ($\alpha=0$, w/o Action KL) baselines. The experimental results show that, with action KL regularization, the token entropy of TARPO follows a controlled decreasing trajectory and converges to a slightly higher level than GRPO.


\paragraph{Soft Token Selection in the Case Study.}
As shown in Figure~\ref{fig:case}, we present a MATH500 case study to illustrate TARPO's soft-token selection during reasoning. Since the selected soft tokens are not human-readable, we display their corresponding hard-token placeholders for visualization. In this case, GRPO makes an incorrect simplification at ``\verb|2(x-5)=4|'', whereas TARPO correctly preserves the cross-multiplication step ``\verb|2(x-5)=4\cdot3|'' and reaches the correct solution. The highlighted tokens further show that GRPO's high-entropy positions are mainly structural transition words such as ``will'' and ``Now we need'', while TARPO assigns higher soft probability to key mathematical tokens, such as equations and operators. 

\section{Conclusion}

In this paper, we presented TARPO, a pure reinforcement learning framework that enables token-wise latent-explicit reasoning through a lightweight action-routing policy. 
Following the GRPO training paradigm, TARPO jointly optimizes the routing head and the LLM backbone with a shared advantage signal, allowing the model to adaptively switch between hard token generation and soft latent reasoning without relying on heuristic rules or supervised initialization. 
Extensive experiments demonstrate that TARPO consistently improves reasoning performance over existing latent reasoning RL methods, while also achieving better token efficiency and strong generalization on out-of-distribution benchmarks.


\section*{Limitations}




While TARPO demonstrates consistent improvements over existing baselines, we discuss two primary limitations of the current study. 

First, regarding the experimental scope, our empirical evaluation is constrained by available computational resources, limiting our exploration to models up to 8B parameters. Although we extensively evaluated TARPO on the Qwen2.5 series and validated cross-architecture generalization on Llama-3.1-8B-Instruct, the largest backbone explored in this work contains 8B parameters. Whether TARPO maintains its effectiveness and training stability at substantially larger scales (e.g., 30B parameters or beyond) remains an open question.

Second, from a methodological perspective, the continuous latent representations in TARPO are constructed via a top-$k$ probability-weighted mixture of token embeddings. While the action head router successfully introduces token-wise structural exploration over reasoning modes (discrete vs.\ continuous), the generation process within the soft mode itself remains deterministic. A promising direction for future work is to integrate TARPO with reparameterization techniques, such as applying the Gumbel-Softmax trick or injecting Gaussian noise into the continuous representations, to complement our structural routing with representation-level exploration.


\bibliography{custom}

\appendix

\newpage 



\section{Implementation Details}
\label{app:setting}



For fair comparison, we follow the training framework and implementation setup of HRPO~\citep{hrpo}, which is built upon TRL and the Unsloth acceleration library\footnote{\url{https://github.com/unslothai/unsloth}}. 
All models are trained for 1 epoch using the 8-bit AdamW optimizer with a cosine learning rate scheduler and a 10\% warmup ratio. All experiments are conducted on a single NVIDIA H800 80GB GPU. Leveraging the Unsloth acceleration framework, training Llama-3.1-8B-Instruct on MATH requires approximately 44 hours.

\begin{table}[htbp]
\centering
\small
\begin{tabular}{lc}
\toprule
\textbf{Hyperparameter} & \textbf{Value} \\
\midrule
Algorithm & TARPO \\
Epochs & 1 \\
Optimizer & AdamW 8bit \\
Optimizer Momentum $\beta_1, \beta_2$ & 0.9, 0.99 \\
Weight Decay & 0.1 \\
Learning Rate (LLM) & 5e-6 / 2e-6 \\
Learning Rate (Action Head) & 1e-4 / 2e-6 \\
TARPO Token KL $\beta$ & 0.005 \\
TARPO Action KL $\alpha$ & 1.0 \\
Action Bias $b_0$ & 2.2 / 4.6 \\
Action Temperature & 1.0 \\
Temperature & 0.5 \\
Group Size $g$ & 4 / 8 \\
Soft Token Top-$k$ & 30 \\
Max Gradient Norm & 0.1 \\
Total Train Batch Size & 32 / 64 \\
LR Scheduler & Cosine with Warmup \\
Warmup Ratio & 0.1 \\
Precision (WA) & BF16-mixed \\
\midrule
\multicolumn{2}{l}{\textbf{LoRA Settings}} \\
\midrule
LoRA Modules & query, key, value, dense \\
LoRA Rank $r$ & 32 \\
LoRA $\alpha$ & 64 \\
\bottomrule
\end{tabular}
\caption{Hyperparameter settings of TARPO.}
\label{tab:all_hyperparameters}
\end{table}

Detailed hyperparameter settings of TARPO are summarized in Table~\ref{tab:all_hyperparameters}.
In addition, several hyperparameters are adjusted according to the dataset and backbone scale. Specifically, the rollout group size $g$ is set to 4 for GSM8K and 8 for MATH and DAPO-MATH-17k. For the Qwen-based models, the backbone learning rate is set to 5e-6, while the action head learning rate is set to 1e-4. For the Llama-3.1-8B-Instruct model, we apply a smaller learning rate of 2e-6 for both the backbone and the action head. The maximum prompt and completion lengths used during training and testing are summarized in Table~\ref{tab:length_settings}. For GSM8K, MATH, and MATH500, we follow the evaluation length settings used in HRPO~\citep{hrpo}. For additional benchmarks not evaluated in HRPO, we adopt task-specific settings. In particular, when evaluating Llama-3.1-8B-Instruct, we set the prompt and completion lengths to 512 and 1024 for GSM8K, and 2048 and 2048 for MATH500.

\begin{table}[htbp]
\centering
\small
\begin{tabular}{llcc}
\toprule
\textbf{Phase} & \textbf{Dataset} & \textbf{Prompt} & \textbf{Completion} \\
\midrule
\multirow{3}{*}{Training} 
 & GSM8K & 1024 & 1024 \\
 & MATH & 1024 & 1024 \\
 & DAPO-MATH-17k & 1024 & 2048 \\
\midrule
\multirow{7}{*}{Testing}
 & GSM8K & 512 & 512 \\
 & MATH & 512 & 1024 \\
 & MATH500 & 512 & 1024 \\
 & AMC23 & 2048 & 2048 \\
 & OlympiadBench & 4096 & 4096 \\
 & GPQA-Diamond & 1024 & 1024 \\
 & ARC-C & 1024 & 1024 \\
 & HumanEval & 2048 & 2048 \\
\bottomrule
\end{tabular}
\caption{Maximum prompt and completion lengths across different datasets for training and testing phases.}
\label{tab:length_settings}
\end{table}

\begin{figure*}[t]
  \centering
  \includegraphics[width=0.9\linewidth]{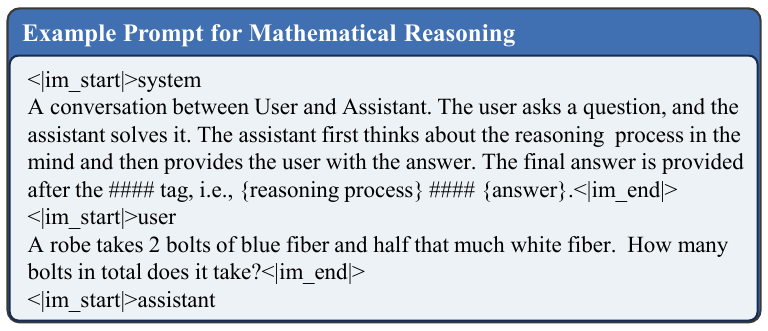}
  \caption{Example Prompt for Mathematical Reasoning in TARPO.}
  \label{fig:prompt1}
\end{figure*}
\begin{figure*}[t]
  \centering
  \includegraphics[width=0.9\linewidth]{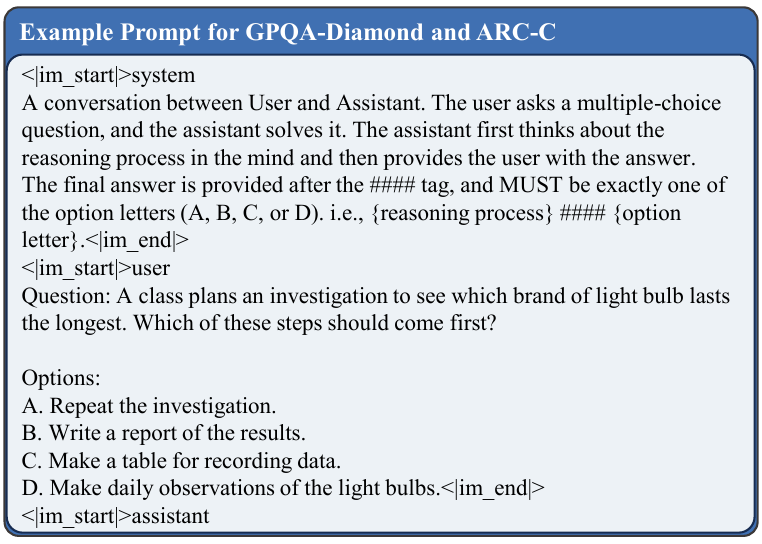}
  \caption{Example Prompt for GPQA-Diamond and ARC-C in TARPO.}
  \label{fig:prompt2}
\end{figure*}
\begin{figure*}[t]
  \centering
  \includegraphics[width=0.9\linewidth]{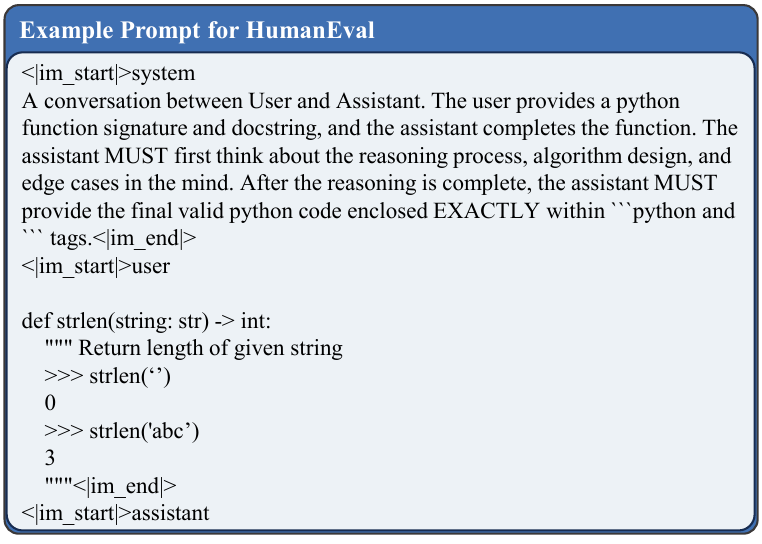}
  \caption{Example Prompt for HumanEval in TARPO.}
  \label{fig:prompt3}
\end{figure*}

We present the exact prompts utilized within the TARPO framework to facilitate reproducibility. As with our experimental configuration, these templates follow HRPO~\citep{hrpo}. Specifically, Figure \ref{fig:prompt1} illustrates the prompt template applied to all mathematical reasoning tasks, including GSM8K, MATH, MATH500, AMC23, OlympiadBench, and DAPO-MATH-17k. Additionally, we provide the specific prompts for our out-of-distribution evaluations, with GPQA-Diamond and ARC-C detailed in Figure \ref{fig:prompt2}, and HumanEval in Figure \ref{fig:prompt3}.

\section{Routing Hyperparameter Analysis}
\label{app:bias_kl}



We conduct a systematic analysis of two routing hyperparameters in TARPO: the initial action-head bias ($b_0$) and the action KL regularization coefficient ($\alpha$).

\begin{figure*}[t]
  \centering
  \begin{subfigure}{0.328\linewidth}
    \includegraphics[width=\linewidth]{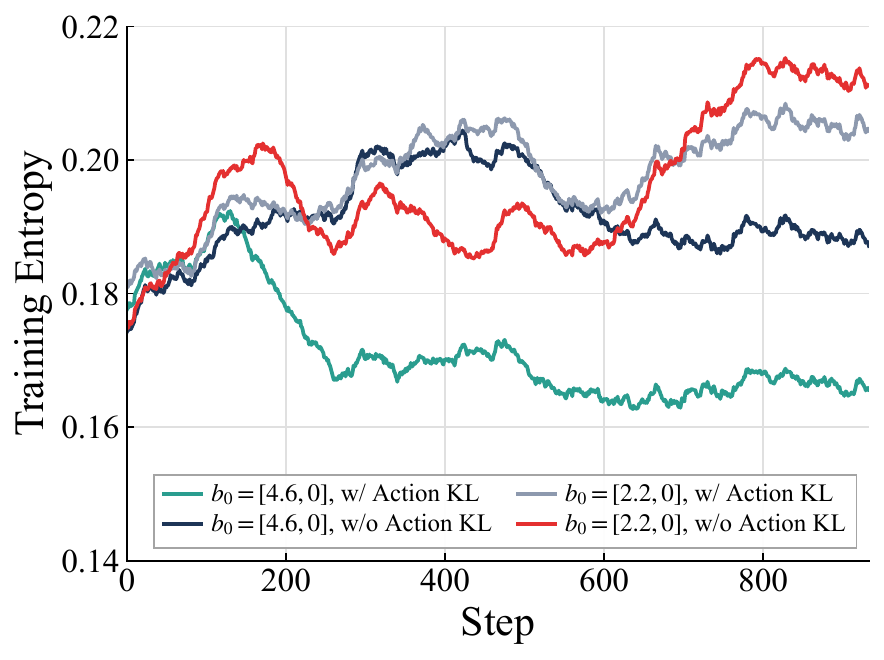}
    \caption{Training Entropy.}
    \label{fig:bias_7b_a}
  \end{subfigure}
  \hfill%
  \begin{subfigure}{0.328\linewidth}
    \includegraphics[width=\linewidth]{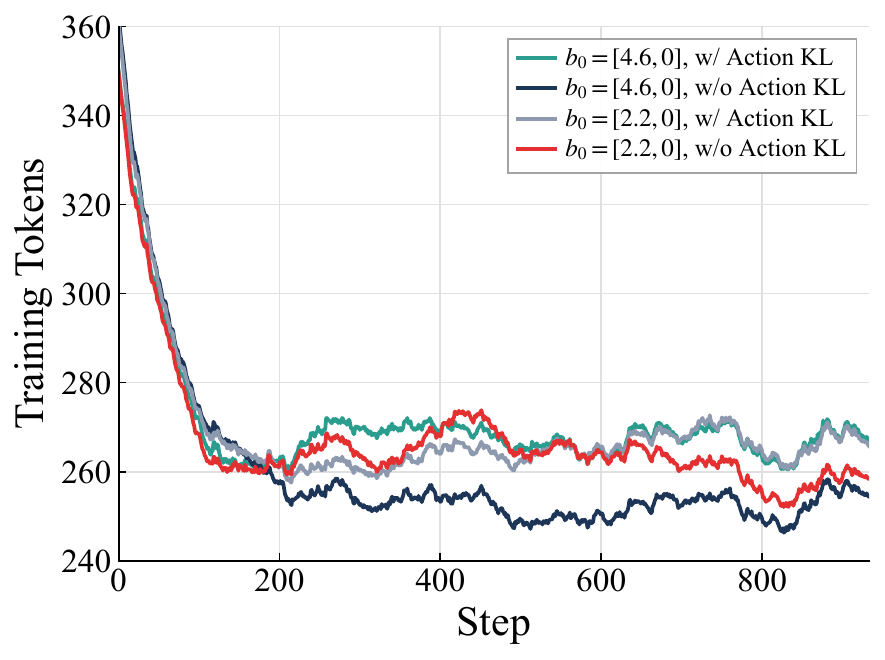}
    \caption{Training Tokens.}
    \label{fig:bias_7b_b}
  \end{subfigure}
  \hfill%
  \begin{subfigure}{0.328\linewidth}
    \includegraphics[width=\linewidth]{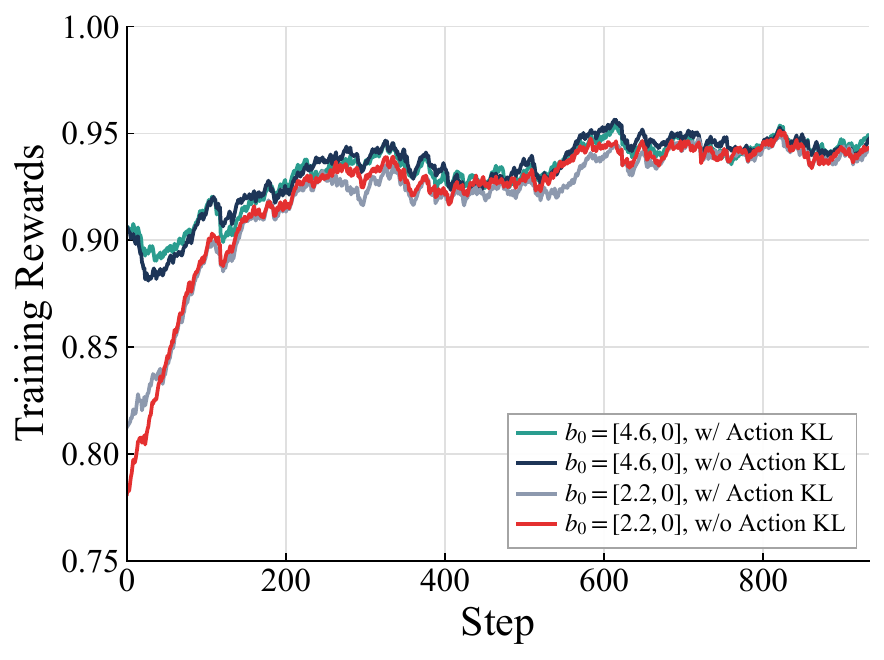}
    \caption{Training Rewards.}
    \label{fig:bias_7b_c}
  \end{subfigure}
 \caption{ Effect of action head initial bias ($b_{0}$) and action KL regularization on Qwen2.5-7B-Instruct trained on GSM8K, under four configurations combining two bias settings ($b_0 = [2.2, 0]$ and $b_0 = [4.6, 0]$) with and without action KL penalty. }
  \label{fig:bias_7b}
\end{figure*}
\begin{figure*}[t]
  \centering
  \begin{subfigure}{0.328\linewidth}
    \includegraphics[width=\linewidth]{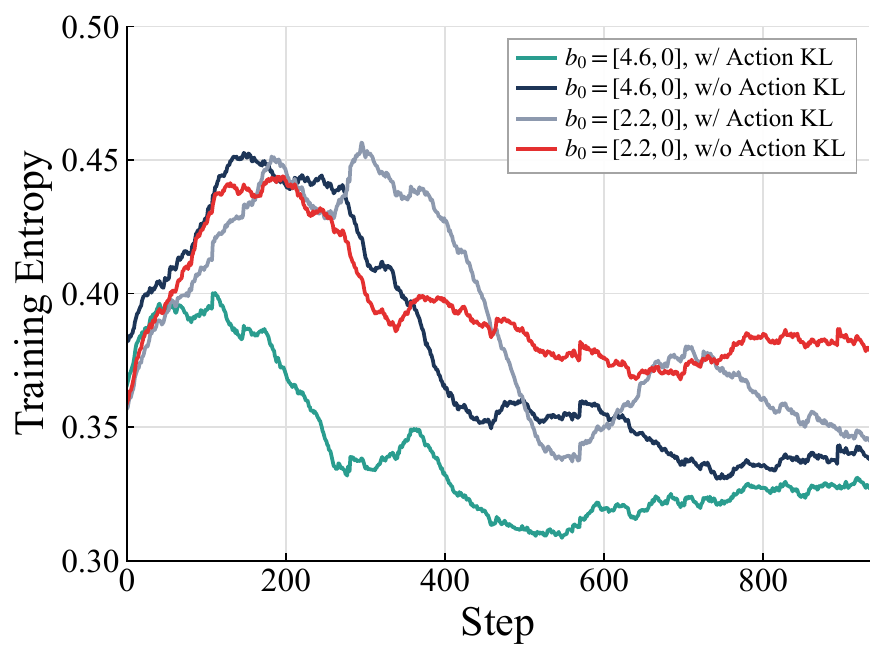}
    \caption{Training Entropy.}
    \label{fig:bias_3b_a}
  \end{subfigure}
  \hfill%
  \begin{subfigure}{0.328\linewidth}
    \includegraphics[width=\linewidth]{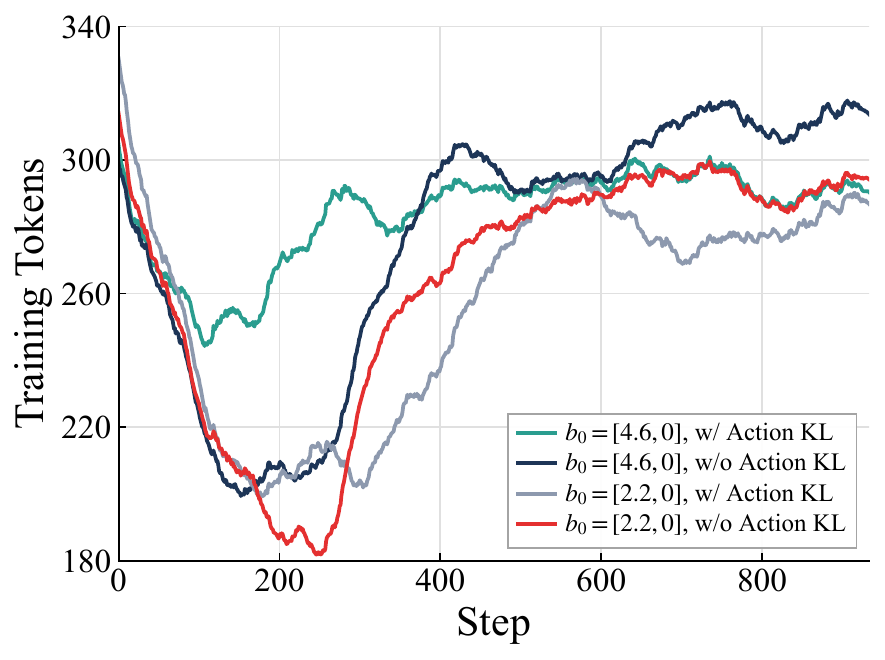}
    \caption{Training Tokens.}
    \label{fig:bias_3b_b}
  \end{subfigure}
  \hfill%
  \begin{subfigure}{0.328\linewidth}
    \includegraphics[width=\linewidth]{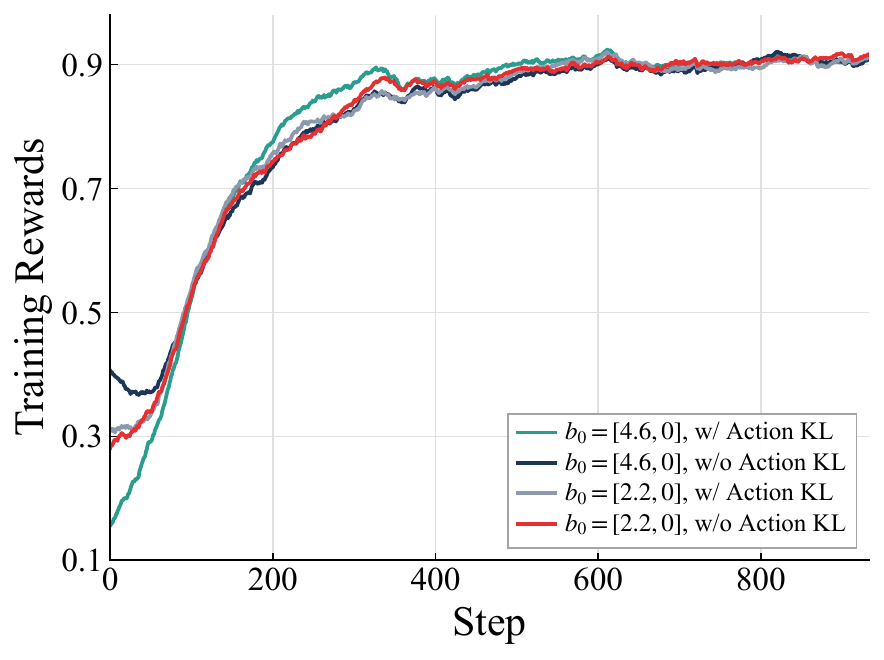}
    \caption{Training Rewards.}
    \label{fig:bias_3b_c}
  \end{subfigure}
 \caption{ Effect of action head initial bias ($b_{0}$) and action KL regularization on Qwen2.5-3B-Instruct trained on GSM8K.}
  \label{fig:bias_3b}
\end{figure*}

Figures~\ref{fig:bias_7b} and~\ref{fig:bias_3b} show the training entropy, token usage, and reward dynamics under different configurations on GSM8K. A stronger hard-routing bias ($b_0 = [4.6, 0]$) results in lower soft-token usage and slower entropy decay, while a weaker bias ($b_0 = [2.2, 0]$) encourages more aggressive latent exploration during early training. Enabling action KL regularization stabilizes the routing dynamics and mitigates premature collapse to degenerate routing behaviors. In contrast, removing action KL leads to more unstable entropy and token-usage trajectories in several settings.

Tables~\ref{tab:bias_kl_3b} and~\ref{tab:bias_kl_id} further show that action KL generally improves training stability and downstream performance, particularly under weaker hard-routing initialization. The 7B model is relatively insensitive to the choice of $b_0$, whereas the 1.5B model exhibits slightly higher sensitivity. Based on these observations, we use $b_0=[2.2,0]$ with action KL for 3B-scale experiments and $b_0=[4.6,0]$ for the remaining scales.

\begin{table*}[t]
\centering
\small 
\setlength{\tabcolsep}{3pt} 
\resizebox{\textwidth}{!}{ 
\begin{tabular}{lc ccc ccc ccc ccc ccc} 
\toprule
\multirow{2}{*}{Bias} & \multirow{2}{*}{Act-KL} & \multicolumn{3}{c}{GSM8K} & \multicolumn{3}{c}{MATH} & \multicolumn{3}{c}{MATH500} & \multicolumn{3}{c}{AMC23} & \multicolumn{3}{c}{OlympiadBench} \\
\cmidrule(lr){3-5} \cmidrule(lr){6-8} \cmidrule(lr){9-11} \cmidrule(lr){12-14} \cmidrule(lr){15-17}
& & P@1 & P@32 & M@32 & P@1 & P@32 & M@32 & P@1 & P@32 & M@32 & P@1 & P@32 & M@32 & P@1 & P@32 & M@32 \\
\midrule
\midrule
\multicolumn{17}{c}{Qwen2.5-3B-Instruct} \\
\midrule
$[4.6, 0]$ & $\times$     & 82.07 & 97.19 & 89.69 & 60.27 & 85.21 & 70.43 & 61.01 & 86.40 & 71.40 & 38.91 & 85.00 & 57.50 & 23.98 & 54.37 & 35.11 \\
$[4.6, 0]$ & $\checkmark$ & 82.91 & 97.19 & 90.22 & 60.06 & 85.10 & 69.96 & 60.51 & 86.20 & 70.80 & 38.98 & 82.50 & 55.00 & 23.85 & 54.67 & 35.56 \\
$[2.2, 0]$ & $\times$     & 82.75 & 97.12 & 89.84 & 59.46 & 85.60 & 70.70 & 60.22 & 86.40 & 69.80 & 38.75 & 85.00 & 57.50 &       &       &       \\
$[2.2, 0]$ & $\checkmark$ & 83.23 & 96.97 & 89.99 & 59.75 & 85.20 & 70.18 & 60.57 & 84.80 & 69.20 & 40.23 & 90.00 & 52.50 & 24.27 & 57.04 & 34.96 \\
\bottomrule
\end{tabular} 
}
\caption{ Ablation of Action Head Bias and Action KL in TARPO on In-Domain Benchmarks. 
}
\label{tab:bias_kl_3b}
\end{table*}

\begin{table*}[t]
\centering
\small 
\setlength{\tabcolsep}{3pt} 
\resizebox{\textwidth}{!}{ 
\begin{tabular}{lc ccc ccc ccc ccc ccc} 
\toprule
\multirow{2}{*}{Bias} & \multirow{2}{*}{Act-KL} & \multicolumn{3}{c}{GSM8K} & \multicolumn{3}{c}{MATH} & \multicolumn{3}{c}{MATH500} & \multicolumn{3}{c}{AMC23} & \multicolumn{3}{c}{OlympiadBench} \\
\cmidrule(lr){3-5} \cmidrule(lr){6-8} \cmidrule(lr){9-11} \cmidrule(lr){12-14} \cmidrule(lr){15-17}
& & P@1 & P@32 & M@32 & P@1 & P@32 & M@32 & P@1 & P@32 & M@32 & P@1 & P@32 & M@32 & P@1 & P@32 & M@32 \\
\midrule
\midrule
\multicolumn{17}{c}{Qwen2.5-7B-Instruct} \\
\midrule
$[4.6, 0]$ & $\checkmark$ & 89.92 & 97.27 & 93.18 & 70.22 & 88.22 & 77.04 & 70.26 & 87.80 & 76.20 & 53.52 & 95.00 & 65.00 & 30.69 & 54.22 & 43.11 \\
$[2.2, 0]$ & $\checkmark$ & 89.82 & 97.50 & 93.18 & 69.93 & 88.28 & 76.98 & 69.91 & 88.20 & 77.60 & 52.03 & 90.00 & 62.50 & 30.56 & 54.22 & 42.07 \\
\midrule
\midrule
\multicolumn{17}{c}{Qwen2.5-1.5B-Instruct} \\
\midrule
$[4.6, 0]$ & $\checkmark$ & 69.96 & 96.59 & 83.09 & 49.71 & 82.28 & 63.04 & 49.71 & 83.40 & 64.20 & 23.98 & 75.00 & 40.00 & 16.13 & 48.89 & 26.07 \\
$[2.2, 0]$ & $\checkmark$ & 68.65 & 96.97 & 82.41 & 47.58 & 81.80 & 61.06 & 48.45 & 83.60 & 61.60 & 22.66 & 72.50 & 37.50 &       &       &       \\
\bottomrule
\end{tabular} 
}
\caption{ In-Domain Ablation of Action Head Bias in TARPO with Action KL Enabled. 
}
\label{tab:bias_kl_id}
\end{table*}

\begin{table*}[t]
\centering
\small 
\setlength{\tabcolsep}{4pt} 
\resizebox{\textwidth}{!}{ 
\begin{tabular}{lc cccc cccc cccc} 
\toprule
\multirow{2}{*}{Bias} & \multirow{2}{*}{Act-KL} & \multicolumn{4}{c}{GPQA-Diamond} & \multicolumn{4}{c}{ARC-C} & \multicolumn{4}{c}{HumanEval} \\
\cmidrule(lr){3-6} \cmidrule(lr){7-10} \cmidrule(lr){11-14} 
& & P@1 & P@32 & M@32 & \#Tok & P@1 & P@32 & M@32 & \#Tok & P@1 & P@32 & M@32 & \#Tok \\
\midrule
\midrule
\multicolumn{14}{c}{Qwen2.5-3B-Instruct} \\
\midrule
$[4.6, 0]$ & $\times$     & 28.09 & 88.38 & 30.30 & 532.3 & 74.01 & 98.89 & 84.04 & 189.4 & 60.25 & 89.63 & 55.49 & 301.7 \\
$[4.6, 0]$ & $\checkmark$ & 28.41 & 90.91 & 29.29 & 568.3 & 74.55 & 97.87 & 85.49 & 213.7 & 63.62 & 88.41 & 61.59 & 256.1 \\
\bottomrule
\end{tabular} 
}
\caption{ OOD effect of Action KL in TARPO under fixed bias initialization.
}
\label{tab:bias_kl_ood}
\end{table*}

\section{Additional Experimental Statistics}
\label{app:exp_stat}


We provide additional statistical results to complement the main evaluation in Section~\ref{sec:experiment}. Specifically, we report Maj@32 (M@32), i.e., the majority-voting accuracy over 32 sampled generations. A question is counted as correct if the most frequent generated answer among the 32 samples matches the ground truth.

In addition, we report the standard deviation (SD) and 95\% confidence intervals (CIs) computed over question-level Avg@32 accuracies, where Avg@32 denotes the fraction of correct samples among 32 generations for each question. The confidence intervals are computed using normal approximation.

\paragraph{In-Domain Benchmarks.}

Tables~\ref{tab:m32_sd} and~\ref{tab:ci} report M@32 together with the SD and 95\% CIs of question-level Avg@32 accuracies across all in-domain reasoning benchmarks for three Qwen2.5 backbones. TARPO consistently achieves M@32 scores comparable to or higher than those of GRPO and HRPO across all scales, while maintaining similar or lower variability in Avg@32 accuracies. The confidence intervals of RL-trained methods (GRPO, HRPO, TARPO) are generally well separated from those of training-free methods (CoT and Entropy-Routed), suggesting that the observed improvements are unlikely to be explained solely by sampling variance.
The Pure Latent method consistently exhibits the highest variability in question-level Avg@32 accuracies across benchmarks. This observation is consistent with our analysis in Section~\ref{sec:ablation}: since stochasticity is concentrated mainly at the final answer-generation stage, the model exhibits larger variance across sampled generations, which also leads to substantially degraded M@32 performance.

\begin{table*}[!t]
\centering
\small
\setlength{\tabcolsep}{3pt}
\resizebox{\textwidth}{!}{
\begin{tabular}{l cc cc cc cc cc}
\toprule
\multirow{2}{*}{Method} & \multicolumn{2}{c}{GSM8K} & \multicolumn{2}{c}{MATH} & \multicolumn{2}{c}{MATH500} & \multicolumn{2}{c}{AMC23} & \multicolumn{2}{c}{Olympiad} \\
\cmidrule(lr){2-3} \cmidrule(lr){4-5} \cmidrule(lr){6-7} \cmidrule(lr){8-9} \cmidrule(lr){10-11}
& M@32 & SD & M@32 & SD & M@32 & SD & M@32 & SD & M@32 & SD \\
\midrule
\midrule
\multicolumn{11}{c}{Qwen2.5-1.5B-Instruct} \\
\midrule
CoT & 74.53 & 0.344 & 51.42 & 0.303 & 49.60 & 0.309 & \textbf{42.50} & 0.218 & 24.59 & 0.241 \\
Pure Latent & 62.02 & 0.470 & 27.96 & 0.396 & 29.20 & 0.406 & 17.50 & 0.353 & 14.22 & 0.311 \\
Entropy-Routed & 68.01 & 0.376 & 40.78 & 0.294 & 41.00 & 0.311 & 30.00 & 0.230 & 22.81 & 0.226 \\
\midrule
GRPO & 83.09 & 0.333 & 61.72 & 0.394 & 63.00 & 0.389 & \textbf{42.50} & 0.296 & 26.52 & 0.277 \\
HRPO & 83.17 & 0.331 & 61.88 & 0.395 & 61.40 & 0.394 & \textbf{42.50} & 0.288 & \textbf{27.70} & 0.280 \\
\textbf{TARPO (Ours)} & \textbf{84.69} & 0.325 & \textbf{63.04} & 0.399 & \textbf{64.20} & 0.397 & \textbf{42.50} & 0.286 & 26.07 & 0.279 \\
\midrule
\midrule
\multicolumn{11}{c}{Qwen2.5-3B-Instruct} \\
\midrule
CoT & 88.40 & 0.288 & 70.40 & 0.383 & 71.00 & 0.383 & 50.00 & 0.349 & 34.67 & 0.335 \\
Pure Latent & 76.27 & 0.414 & 57.00 & 0.482 & 57.40 & 0.483 & 45.00 & 0.487 & 28.74 & 0.424 \\
Entropy-Routed & 81.43 & 0.329 & 66.42 & 0.397 & 67.00 & 0.401 & \textbf{52.50} & 0.348 & 33.48 & 0.340 \\
\midrule
GRPO & \textbf{90.22} & 0.285 & 71.10 & 0.403 & \textbf{71.20} & 0.404 & \textbf{52.50} & 0.354 & 35.41 & 0.341 \\
HRPO & 89.76 & 0.286 & \textbf{70.70} & 0.403 & 69.80 & 0.401 & \textbf{52.50} & 0.352 & \textbf{37.04} & 0.342 \\
\textbf{TARPO (Ours)} & 89.99 & 0.286 & 70.18 & 0.403 & 69.20 & 0.406 & \textbf{52.50} & 0.357 & 34.96 & 0.345 \\
\midrule
\midrule
\multicolumn{11}{c}{Qwen2.5-7B-Instruct} \\
\midrule
CoT & 92.65 & 0.243 & 74.44 & 0.376 & 73.80 & 0.379 & 62.50 & 0.349 & \textbf{43.85} & 0.331 \\
Pure Latent & 89.92 & 0.307 & 66.12 & 0.444 & 68.20 & 0.440 & 60.00 & 0.453 & 35.85 & 0.421 \\
Entropy-Routed & 91.58 & 0.259 & 67.50 & 0.390 & 65.35 & 0.399 & 57.50 & 0.358 & 39.85 & 0.352 \\
\midrule
GRPO & \textbf{93.33} & 0.240 & 76.94 & 0.391 & 76.00 & 0.394 & 62.50 & 0.383 & 42.96 & 0.398 \\
HRPO & 93.18 & 0.238 & 76.74 & 0.391 & 76.00 & 0.395 & 62.50 & 0.389 & 42.22 & 0.397 \\
\textbf{TARPO (Ours)} & 93.03 & 0.241 & \textbf{77.04}& 0.391 & \textbf{76.20}& 0.393 & \textbf{65.00} & 0.370 & 43.11 & 0.393 \\
\bottomrule
\end{tabular}
}
\caption{Detailed performance statistics including Maj@32 (M@32, majority voting accuracy over 32 samples) and Standard Deviation (SD) on in-domain reasoning benchmarks across the Qwen2.5-Instruct model family (1.5B, 3B, and 7B). The best Maj@32 results for each model scale are highlighted in bold. 
}
\label{tab:m32_sd}
\end{table*}

\begin{table*}[!t]
\centering
\small
\setlength{\tabcolsep}{3pt}
\resizebox{\textwidth}{!}{
\begin{tabular}{l ccccc}
\toprule
Method & GSM8K & MATH & MATH500 & AMC23 & Olympiad \\
\midrule
\midrule
\multicolumn{6}{c}{Qwen2.5-1.5B-Instruct} \\
\midrule
CoT & [59.03, 62.74] & [30.82, 32.50] & [29.05, 34.46] & [12.09, 25.57] & [12.31, 15.95] \\
Pure Latent & [57.99, 63.06] & [24.20, 26.39] & [22.95, 30.07] & [4.93, 26.79] & [9.86, 14.56] \\
Entropy-Routed & [58.63, 62.69] & [25.29, 26.92] & [24.37, 29.83] & [6.86, 21.10] & [10.95, 14.36] \\
\midrule
GRPO & [67.24, 70.83] & [46.73, 48.92] & [45.16, 51.97] & [17.61, 35.98] & [14.33, 18.50] \\
HRPO & [67.93, 71.50] & [47.89, 50.09] & [46.35, 53.25] & [16.85, 34.72] & [14.40, 18.63] \\
\textbf{TARPO (Ours)} & [69.01, 72.52] & [48.61, 50.82] & [46.23, 53.18] & [15.11, 32.86] & [14.03, 18.23] \\
\midrule
\midrule
\multicolumn{6}{c}{Qwen2.5-3B-Instruct} \\
\midrule
CoT & [71.94, 75.05] & [54.58, 56.70] & [52.66, 59.37] & [25.67, 47.30] & [20.30, 25.35] \\
Pure Latent & [71.61, 76.08] & [54.11, 56.79] & [52.46, 60.94] & [28.11, 58.30] & [21.65, 28.04] \\
Entropy-Routed & [72.13, 75.69] & [53.65, 55.85] & [52.06, 59.09] & [26.80, 48.36] & [20.43, 25.55] \\
\midrule
GRPO & [81.01, 84.09] & [58.74, 60.97] & [57.31, 64.39] & [27.70, 49.65] & [21.01, 26.15] \\
HRPO & [81.20, 84.29] & [58.86, 61.09] & [57.12, 64.15] & [27.93, 49.72] & [21.22, 26.37] \\
\textbf{TARPO (Ours)} & [81.69, 84.78] & [58.64, 60.87] & [57.01, 64.13] & [29.16, 51.31] & [21.67, 26.87] \\
\midrule
\midrule
\multicolumn{6}{c}{Qwen2.5-7B-Instruct} \\
\midrule
CoT & [86.66, 89.28] & [60.12, 62.20] & [58.19, 64.84] & [33.41, 55.03] & [23.24, 28.23] \\
Pure Latent & [86.75, 90.05] & [55.55, 58.01] & [55.64, 63.36] & [34.48, 62.55] & [24.38, 30.73] \\
Entropy-Routed & [86.67, 89.46] & [52.94, 55.10] & [50.43, 57.39] & [30.47, 52.65] & [23.92, 29.24] \\
\midrule
GRPO & [88.57, 91.17] & [68.96, 71.13] & [66.40, 73.31] & [39.84, 63.60] & [27.96, 33.97] \\
HRPO & [88.85, 91.42] & [69.17, 71.33] & [66.32, 73.24] & [39.11, 63.23] & [27.75, 33.74] \\
\textbf{TARPO (Ours)} & [88.64, 91.24] & [69.14, 71.30] & [66.81, 73.71] & [42.05, 64.98] & [27.73, 33.65] \\
\bottomrule
\end{tabular}
}
\caption{95\% Confidence Intervals (CI) for Pass@1 accuracy on in-domain reasoning benchmarks across the Qwen2.5-Instruct model family (1.5B, 3B, and 7B).}
\label{tab:ci}
\end{table*}

\paragraph{Out-of-Distribution Benchmarks.}


Table~\ref{tab:ood_stats} extends the OOD evaluation in Table~\ref{tab:ood_results} by reporting M@32 together with the SD and 95\% CIs of Avg@32 accuracies on GPQA-Diamond, ARC-C, and HumanEval for Qwen2.5-3B-Instruct. On the HumanEval benchmark, TARPO achieves the highest M@32 of 61.59, indicating that adaptive routing also performs well under majority-voting evaluation on code generation tasks.

\begin{table*}[t]
\centering
\small 
\setlength{\tabcolsep}{4pt} 
\resizebox{\textwidth}{!}{ 
\begin{tabular}{l ccc ccc ccc} 
\toprule
\multirow{2}{*}{Method} & \multicolumn{3}{c}{GPQA-Diamond} & \multicolumn{3}{c}{ARC-C} & \multicolumn{3}{c}{HumanEval} \\
\cmidrule(lr){2-4} \cmidrule(lr){5-7} \cmidrule(lr){8-10} 
& M@32 & SD & 95\% CI & M@32 & SD & 95\% CI & M@32 & SD & 95\% CI \\
\midrule
\midrule
\multicolumn{10}{c}{Qwen2.5-3B-Instruct} \\
\midrule
CoT            & 28.79 & 0.216 & [23.83, 29.84] & \textbf{86.26} & 0.270 & [72.50, 75.60] & 52.44 & 0.342 & [0.536, 0.641] \\
Pure Latent    & 28.28 & 0.413 & [21.50, 33.01] & 72.95 & 0.426 & [68.13, 73.01] & 56.71 & 0.377 & [54.43, 65.96] \\
Entropy-Routed & \textbf{31.82} & 0.259 & [24.70, 31.90] & 79.69 & 0.319 & [69.26, 72.92] & 54.27 & 0.357 & [52.85, 63.77] \\
\midrule
GRPO           & 28.79 & 0.265 & [24.33, 31.73] & 82.94 & 0.290 & [73.12, 76.45] & 50.00 & 0.318 & [53.99, 63.73] \\
HRPO           & 27.78 & 0.248 & [24.05, 30.97] & 83.62 & 0.284 & [73.43, 76.68] & 58.54 & 0.332 & [54.43, 64.59] \\
\textbf{TARPO} & 29.29 & 0.252 & [24.90, 31.92] & 85.49 & 0.274 & [72.99, 76.12] & \textbf{61.59} & 0.351 & [58.26, 68.99] \\
\bottomrule
\end{tabular} 
}
\caption{ Out-of-distribution (OOD) evaluation results for Qwen2.5-3B-Instruct on GPQA-Diamond, ARC-C, and HumanEval, reporting Maj@32 (M@32), Standard Deviation (SD), and 95\% Confidence Interval (CI). The best overall Maj@32 results are highlighted in bold. }
\label{tab:ood_stats}
\end{table*}

\paragraph{Cross-Architecture Generalization.}


Table~\ref{tab:llama_stats} reports M@32 together with the SD and 95\% CIs of Avg@32 accuracies for Llama-3.1-8B-Instruct on GSM8K, MATH500, and AMC23. TARPO outperforms GRPO in M@32 across all three benchmarks. Moreover, the confidence intervals on MATH500 and AMC23 are well separated, supporting the cross-architecture robustness conclusion discussed in Section~\ref{sec:cross-architecture-results}.

\begin{table*}[t]
\centering
\small 
\setlength{\tabcolsep}{4pt} 
\resizebox{\textwidth}{!}{ 
\begin{tabular}{l ccc ccc ccc} 
\toprule
\multirow{2}{*}{Method} & \multicolumn{3}{c}{GSM8K} & \multicolumn{3}{c}{MATH500} & \multicolumn{3}{c}{AMC23} \\
\cmidrule(lr){2-4} \cmidrule(lr){5-7} \cmidrule(lr){8-10} 
& M@32 & SD & 95\% CI & M@32 & SD & 95\% CI & M@32 & SD & 95\% CI \\
\midrule
\midrule
\multicolumn{10}{c}{Llama-3.1-8B-Instruct} \\
\midrule
CoT            & 89.16 & 0.282 & [76.26, 79.30] & 53.80 & 0.334 & [32.55, 38.40] & 37.50 & 0.216 & [9.48, 22.86] \\
GRPO           & 88.93 & 0.286 & [81.44, 84.53] & 59.80 & 0.391 & [42.35, 49.20] & 35.00 & 0.298 & [12.32, 30.80] \\
\textbf{TARPO} & \textbf{89.76} & 0.284 & [82.72, 85.78] & \textbf{61.40} & 0.395 & [44.19, 51.11] & \textbf{42.50} & 0.316 & [13.19, 32.75] \\
\bottomrule
\end{tabular} 
}
\caption{Cross-architecture generalization results on Llama-3.1-8B-Instruct, reporting Maj@32 (M@32), Standard Deviation (SD), and 95\% Confidence Interval (CI) on GSM8K, MATH500, and AMC23.}
\label{tab:llama_stats}
\end{table*}

\section{Additional Training Curves}
\label{app:curves}

\begin{figure*}[t]
  \centering
  \begin{subfigure}{0.328\linewidth}
    \includegraphics[width=\linewidth]{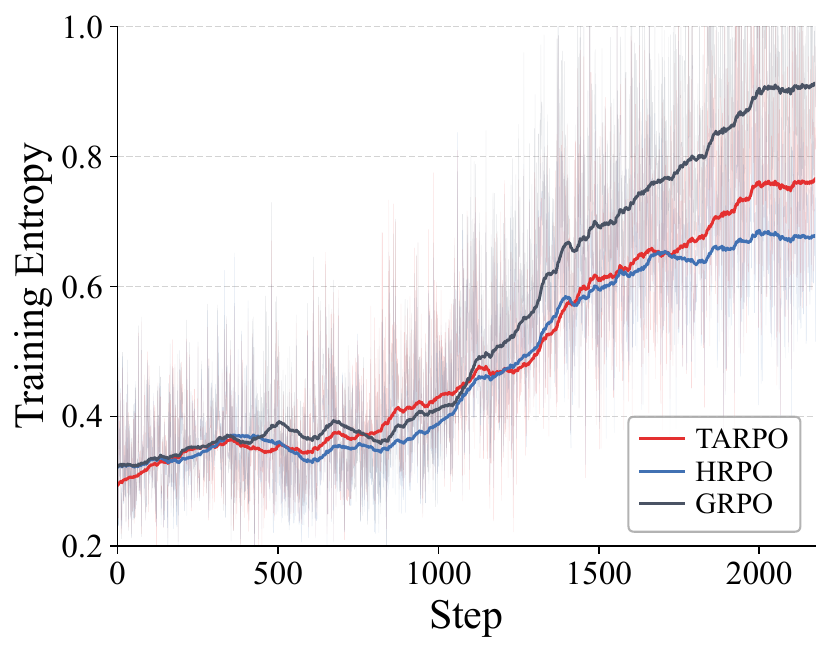}
    \caption{Training Entropy.}
    \label{fig:3b_dapo_a}
  \end{subfigure}
  \hfill%
  \begin{subfigure}{0.328\linewidth}
    \includegraphics[width=\linewidth]{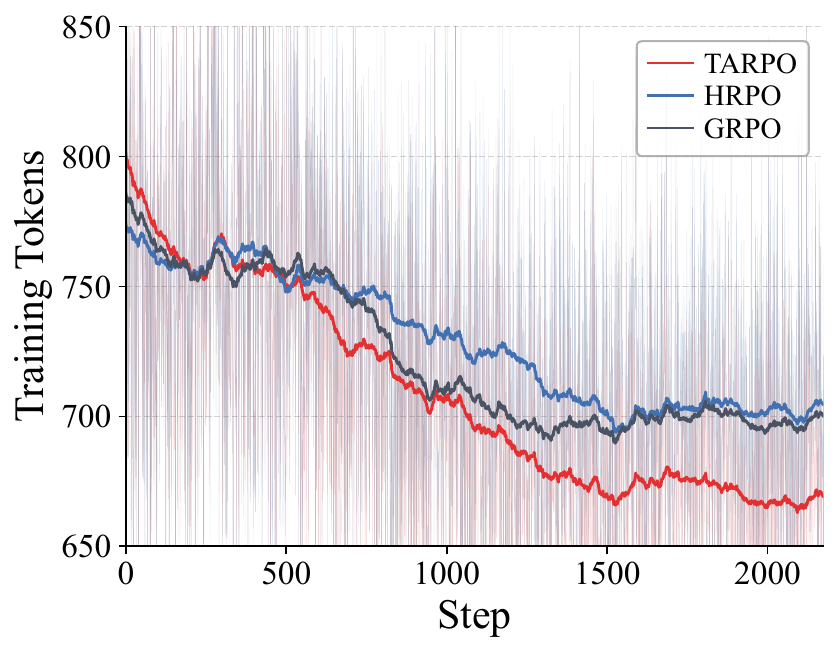}
    \caption{Training Tokens.}
    \label{fig:3b_dapo_b}
  \end{subfigure}
  \hfill%
  \begin{subfigure}{0.328\linewidth}
    \includegraphics[width=\linewidth]{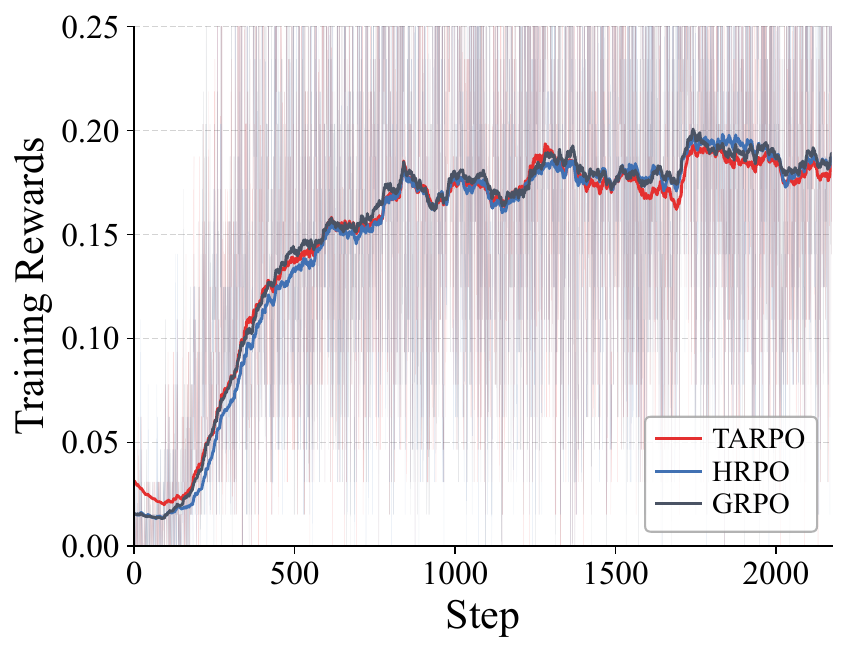}
    \caption{Training Rewards.}
    \label{fig:3b_dapo_c}
  \end{subfigure}
 \caption{ Training curves of Qwen2.5-3B-Instruct on DAPO-MATH-17k.  From left to right: Training Entropy, Training Tokens (counting both hard and soft tokens for TARPO), and Training Rewards. }
  \label{fig:3b_dapo}
\end{figure*}


To complement the training dynamics presented in Section~\ref{sec:training_dynamics}, we provide additional training curves for the backbone and dataset combinations not shown in the main paper. All figures report training entropy, training tokens, and training rewards across optimization steps. Training tokens include both discrete and latent tokens in TARPO.

\begin{figure*}[t]
  \centering
  \begin{subfigure}{0.328\linewidth}
    \includegraphics[width=\linewidth]{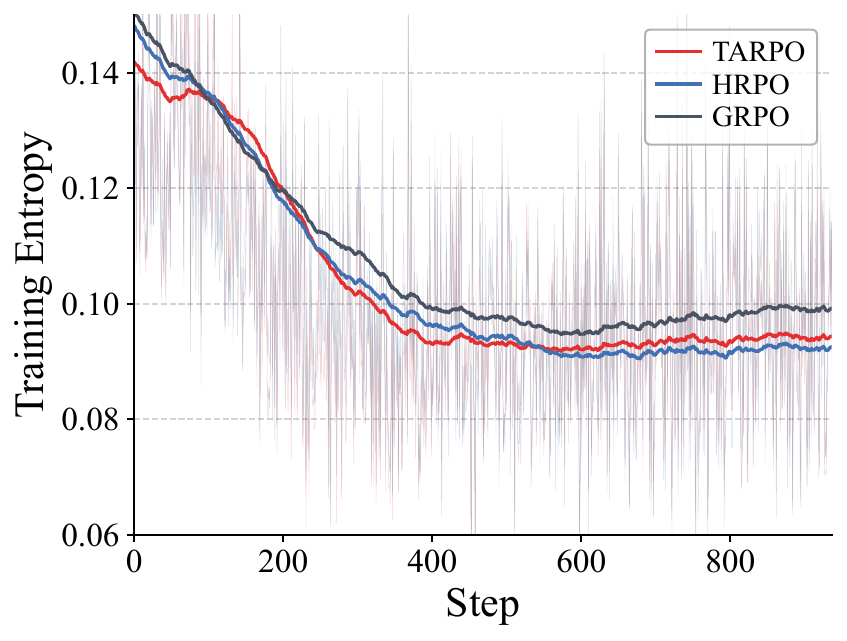}
    \caption{Training Entropy.}
    \label{fig:7b_math_a}
  \end{subfigure}
  \hfill%
  \begin{subfigure}{0.328\linewidth}
    \includegraphics[width=\linewidth]{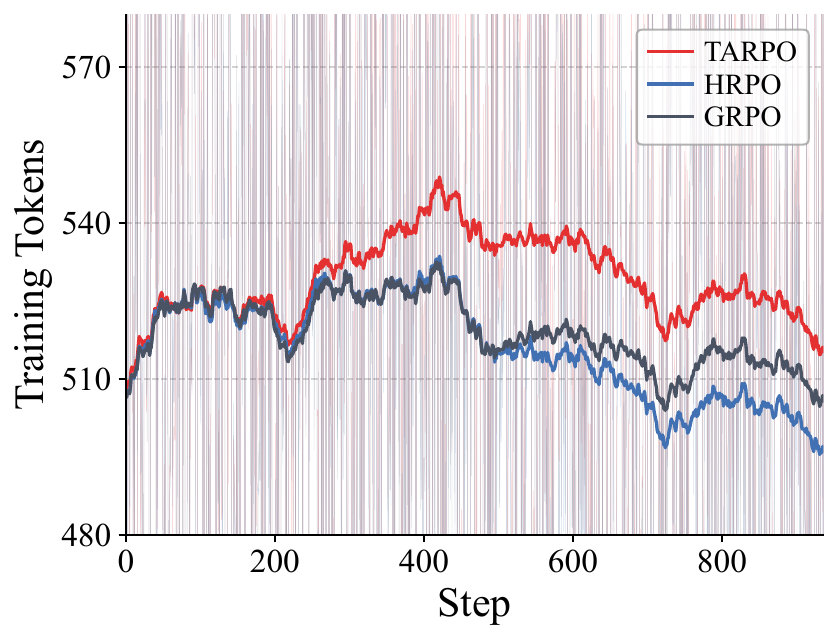}
    \caption{Training Tokens.}
    \label{fig:7b_math_b}
  \end{subfigure}
  \hfill%
  \begin{subfigure}{0.328\linewidth}
    \includegraphics[width=\linewidth]{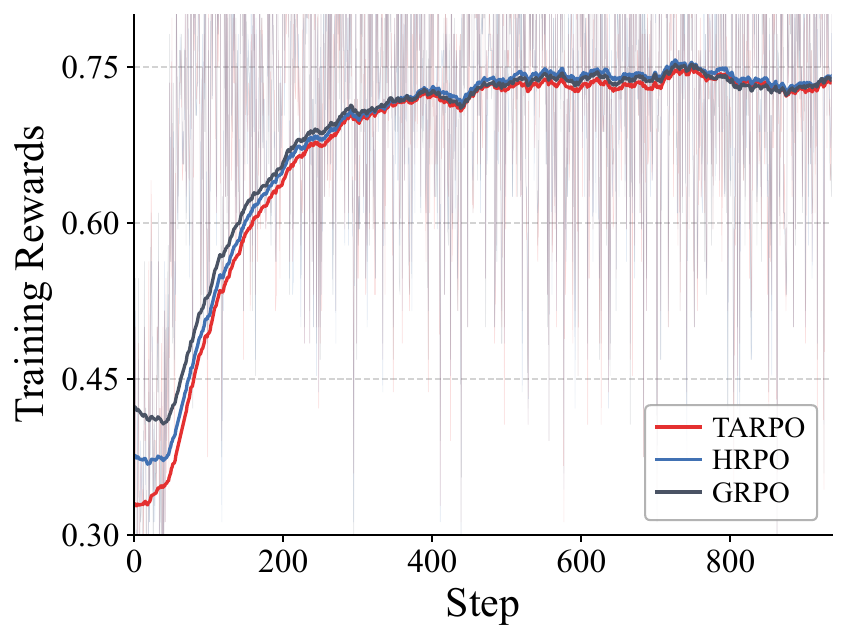}
    \caption{Training Rewards.}
    \label{fig:7b_math_c}
  \end{subfigure}
 \caption{ Training curves of Qwen2.5-7B-Instruct on MATH.}
  \label{fig:7b_math}
\end{figure*}
\begin{figure*}[t]
  \centering
  \begin{subfigure}{0.328\linewidth}
    \includegraphics[width=\linewidth]{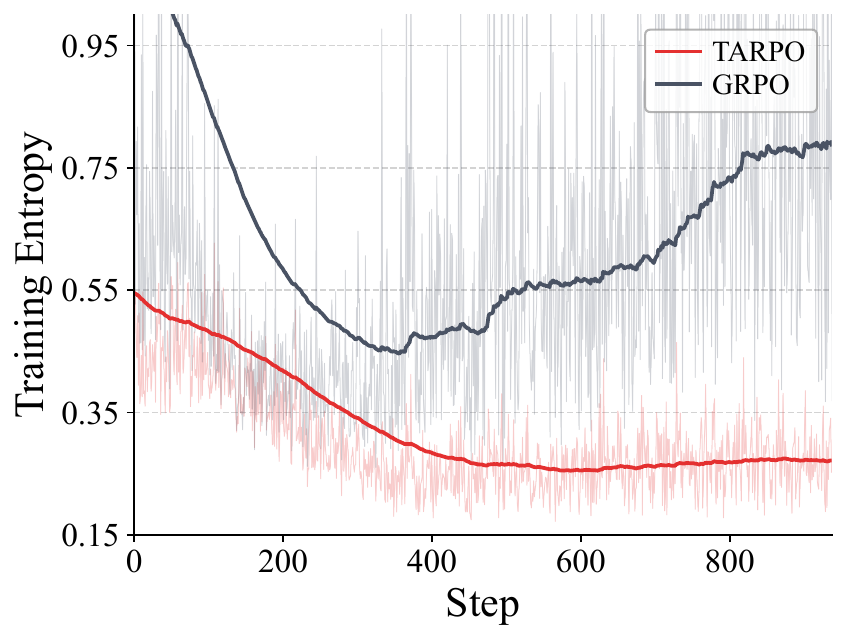}
    \caption{Training Entropy.}
    \label{fig:llama_math_a}
  \end{subfigure}
  \hfill%
  \begin{subfigure}{0.328\linewidth}
    \includegraphics[width=\linewidth]{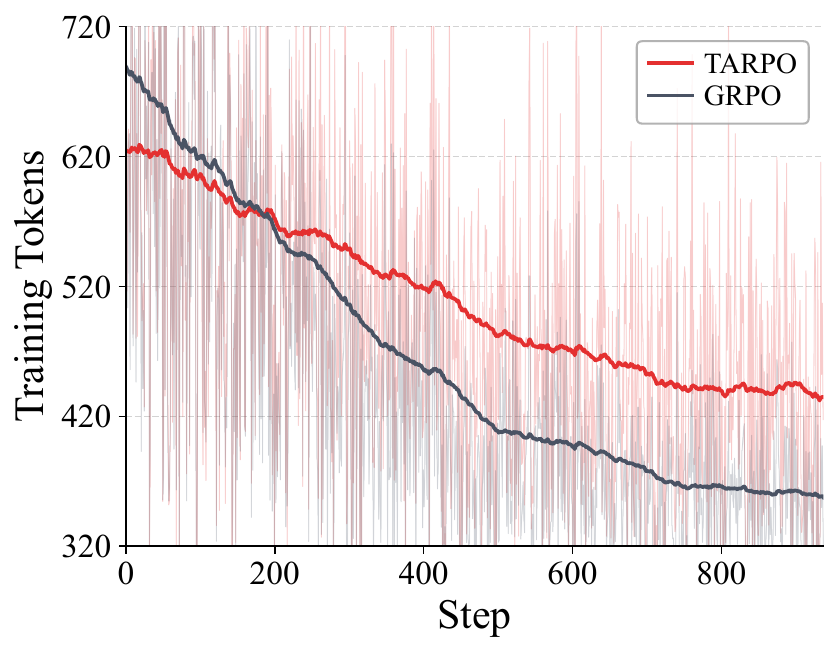}
    \caption{Training Tokens.}
    \label{fig:llama_math_b}
  \end{subfigure}
  \hfill%
  \begin{subfigure}{0.328\linewidth}
    \includegraphics[width=\linewidth]{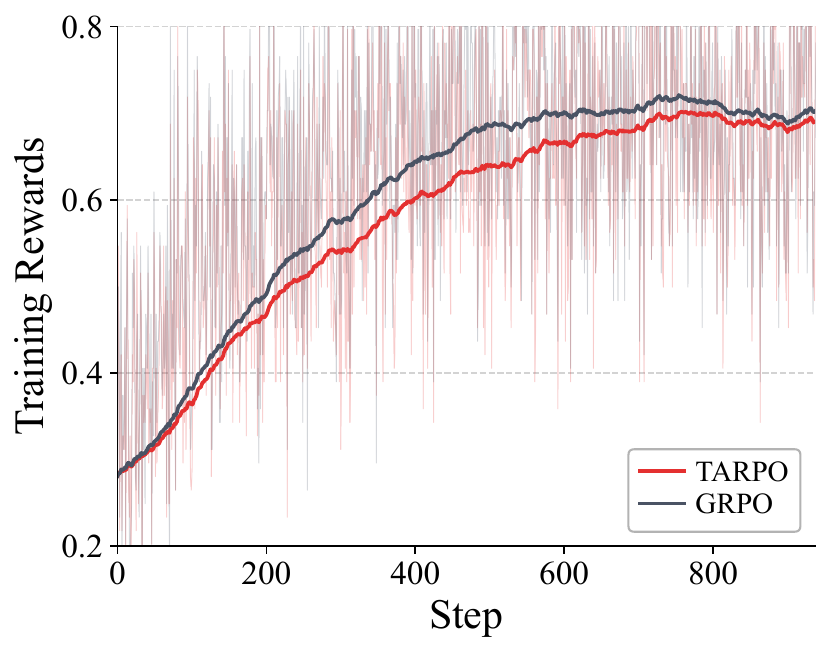}
    \caption{Training Rewards.}
    \label{fig:llama_math_c}
  \end{subfigure}
 \caption{ Training curves of Llama-3.1-8B-Instruct on MATH. }
  \label{fig:llama_math}
\end{figure*}

As shown in Figures~\ref{fig:3b_dapo},~\ref{fig:7b_math}, and~\ref{fig:llama_math}, TARPO exhibits consistent training behavior across these remaining backbone-dataset combinations. On Qwen2.5-7B-Instruct trained on MATH, TARPO consistently produces the shortest completion lengths throughout training while maintaining rewards comparable to HRPO and GRPO. This is consistent with the token-efficiency advantage observed at the 1.5B scale in the main paper. The entropy trajectories of the three methods also remain closely aligned, suggesting that the action-routing mechanism does not destabilize optimization at larger model scales. 
On Qwen2.5-3B-Instruct trained on DAPO-MATH-17k, TARPO similarly achieves competitive reward convergence with lower token usage than the baselines. On Llama-3.1-8B-Instruct trained on MATH, TARPO maintains higher training entropy than GRPO while achieving higher rewards and shorter completions at convergence, suggesting that the routing mechanism preserves effective exploration during training.





\section{Licenses}

For base LLMs, datasets, and frameworks, we list their licenses in Table~\ref{tab:licenses}.

\begin{table*}[h]
\centering
\caption{A summary of licenses.}
\label{tab:licenses}
\resizebox{\textwidth}{!}{%
\begin{tabular}{llll}
\toprule
\textbf{Resources} & \textbf{Type} & \textbf{License} & \textbf{URL} \\
\midrule
Qwen2.5-1.5B-Instruct & Base LLM & Apache-2.0 & \url{https://huggingface.co/Qwen/Qwen2.5-1.5B-Instruct} \\
Qwen2.5-3B-Instruct   & Base LLM & Qwen-Research License & \url{https://huggingface.co/Qwen/Qwen2.5-3B-Instruct} \\
Qwen2.5-7B-Instruct   & Base LLM & Apache-2.0 & \url{https://huggingface.co/Qwen/Qwen2.5-7B-Instruct} \\
Llama-3.1-8B-Instruct & Base LLM & Llama 3.1 Community License  Agreement & \url{https://huggingface.co/meta-llama/Llama-3.1-8B-Instruct} \\
\midrule
HRPO    & RL-framework       & Not Specified        & \url{https://github.com/Yueeeeeeee/HRPO} \\
TRL      & RL-framework       & Apache-2.0 & \url{https://github.com/huggingface/trl} \\
Unsloth  & Training-framework & Apache-2.0, AGPL-3.0 & \url{https://github.com/unslothai/unsloth} \\
\midrule
GSM8K          & Dataset & MIT              & \url{https://github.com/openai/grade-school-math} \\
MATH           & Dataset & MIT              & \url{https://github.com/hendrycks/math} \\
MATH500        & Dataset & MIT              & \url{https://github.com/hendrycks/math} \\
DAPO-MATH-17k  & Dataset & Not Specified & \url{https://huggingface.co/datasets/open-r1/DAPO-Math-17k-Processed} \\
AMC23          & Dataset & Available Online & \url{https://github.com/eric-ai-lab/Soft-Thinking} \\
OlympiadBench  & Dataset & MIT              & \url{https://github.com/zz1358m/SofT-GRPO-master} \\
GPQA-Diamond   & Dataset & MIT              & \url{https://github.com/eric-ai-lab/Soft-Thinking} \\
ARC-C          & Dataset & Apache-2.0       & \url{https://github.com/eric-ai-lab/Soft-Thinking} \\
HumanEval      & Dataset & MIT              & \url{https://github.com/eric-ai-lab/Soft-Thinking} \\
\bottomrule
\end{tabular}%
}
\end{table*}

\end{document}